%% file: template.tex
\documentclass{article}

\PassOptionsToPackage{numbers,compress}{natbib}
\usepackage[final]{neurips_2024}

\usepackage{amsfonts}       
\usepackage{amsmath}        
\usepackage{amssymb}        
\usepackage{array}          
\usepackage{booktabs}       
\usepackage{caption}        
\usepackage{enumitem}       
\usepackage[T1]{fontenc}    
\usepackage{graphicx}       
\usepackage{hhline}         
\usepackage{hyperref}       
\usepackage{cleveref}       
\usepackage[utf8]{inputenc} 
\usepackage{lipsum}         
\usepackage{makecell}       
\usepackage{microtype}      
\usepackage{multicol}       
\usepackage{multirow}       
\usepackage{nicefrac}       
\usepackage{pifont}         
\usepackage{setspace}       
\usepackage{subcaption}     
\usepackage{tabularx}       
\usepackage{tikz}           
\usepackage{url}            
\usepackage[dvipsnames,table]{xcolor} 
\usepackage{enumitem}

\usepackage{tocloft}
\usepackage{hyperref}
\hypersetup{colorlinks=true, linkcolor=brainblue, citecolor=brainblue, urlcolor=brainblue, pdfborder={0 0 0}}

\usepackage{xcolor}
\definecolor{brainblue}{HTML}{2878B5}
\makeatletter
\renewcommand{\@seccntformat}[1]{%
  \textcolor{brainblue}{\csname the#1\endcsname}\quad}
\makeatother

\addtocontents{toc}{\vspace*{5pt}}

\makeatletter

\newlength{\absleft}
\newlength{\absright}
\newcommand{\abodyfont}{\normalsize}     
\newcommand{\atitlefont}{\Large\bfseries} 

\makeatother

\graphicspath{ {./images/} }

\newcommand{\downtriangle}{%
  \tikz[baseline=-0.5ex,  xscale=1.5, yscale=1.8] \filldraw[black] (0,0) -- (1ex,0) -- (0.5ex,-0.8ex) -- cycle;
}
\title{BLM$_1$: A Boundless Large Model for Cross-Space, Cross-Task, and Cross-Embodiment Learning}
\author{} 

\begin{document}
\maketitle

\vspace{-7em} 
\begin{center}
\textbf{Wentao Tan}\textsuperscript{\dag}   \quad 
\textbf{Bowen Wang}   \quad 
\textbf{Heng Zhi}     \quad 
\textbf{Chenyu Liu}   \quad 
\textbf{Zhe Li}       \quad 
\textbf{Jian Liu}     \quad 
\textbf{Zengrong Lin} \quad 
\textbf{Yukun Dai}    \quad 
\textbf{Yipeng Chen}  \quad 
\textbf{Wenjie Yang}  \quad 
\textbf{Enci Xie}     \quad 
\textbf{Hao Xue}      \quad 
\textbf{Baixu Ji}     \quad 
\textbf{Chen Xu}      \quad 
\textbf{Zhibin Wang}  \quad 
\textbf{Tianshi Wang} \quad 
\textbf{Lei Zhu}\textsuperscript{\dag}\textsuperscript{$\diamond$} \quad
\textbf{Heng Tao Shen}\textsuperscript{$\diamond$}
\\[0.35em]
\textit{School of Computer Science and Technology, Tongji University;}\,
\textit{Shanghai Magic;}\,
\textit{Koala Uran}
\\[0.35em]
\url{https://boundless-large-model.github.io}
\\[0.25em]
\textsuperscript{\dag}Project Leader \quad
\textsuperscript{$\diamond$}Corresponding Author
\end{center}

\vspace{-1em} 
\begin{center}
    \includegraphics[width=0.99\textwidth]{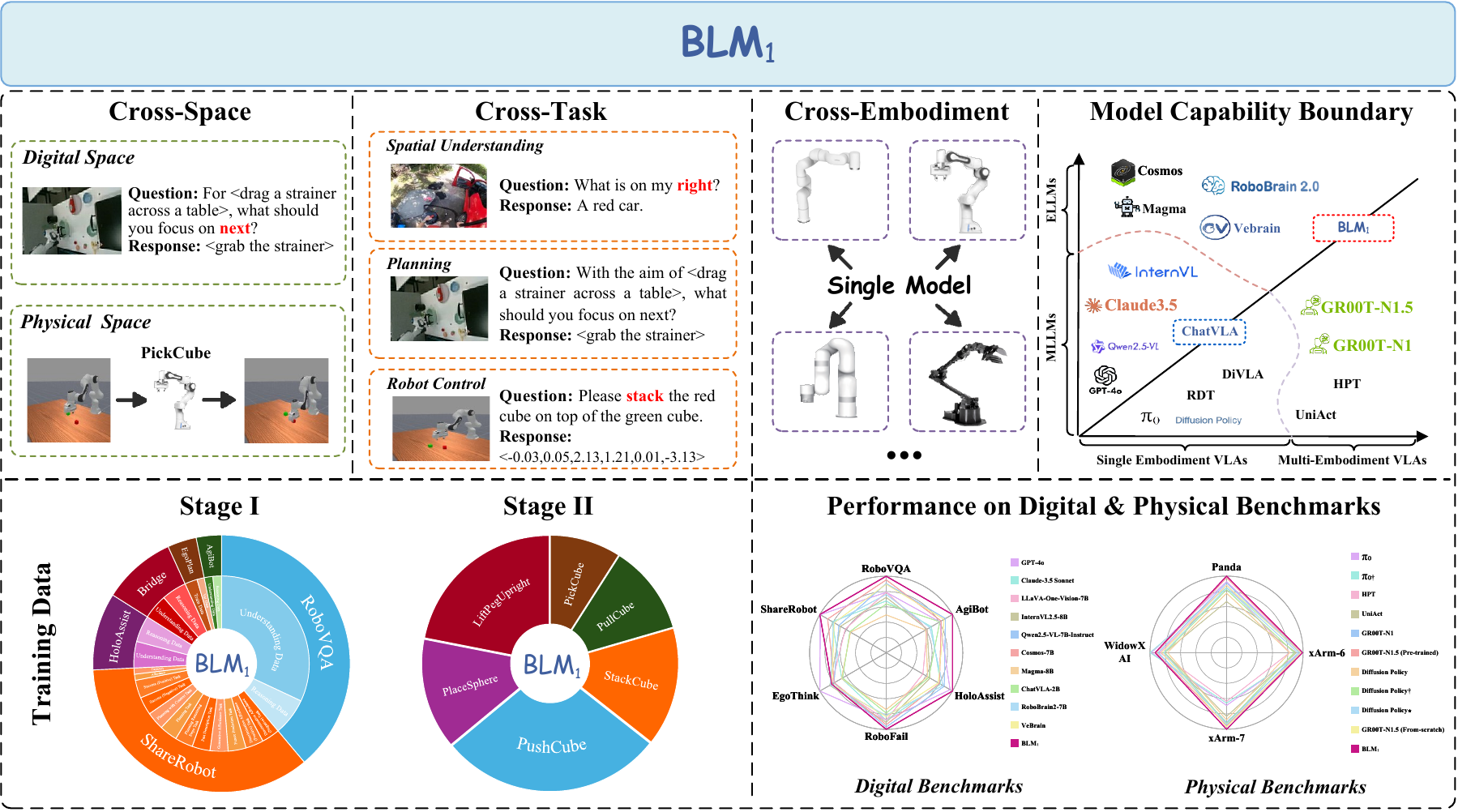}
    \captionof{figure}{BLM$_1$ is the first work to realize cross-space transfer, cross-task learning, and cross-embodiment generalization within a \textit{single} multimodal spatial foundation. Evaluations show BLM$_1$ achieves SOTA performance over MLLMs, ELLMs, VLAs and GMLMs across digital and physical spaces.} 
    \label{fig:system}
\end{center}
\vspace{-1.5em} 

\begin{center}
\Large\textbf{Abstract}
\end{center}
\vspace{-0.5em}

Multimodal large language models (MLLMs) have advanced vision-language reasoning and are increasingly deployed in embodied agents. However, significant limitations remain: MLLMs generalize poorly across digital-physical spaces and embodiments; vision-language-action models (VLAs) produce low-level actions yet lack robust high-level embodied reasoning; and most embodied large language models (ELLMs) are constrained to digital-space with poor generalization to the physical world. Thus, unified models that operate seamlessly across digital and physical spaces while generalizing across embodiments and tasks remain absent. We introduce the \textbf{Boundless Large Model (BLM$_1$)}, a multimodal spatial foundation model that preserves instruction following and reasoning, incorporates embodied knowledge, and supports robust cross-embodiment control. BLM$_1$ integrates three key capabilities---\textit{cross-space transfer, cross-task learning, and cross-embodiment generalization}---via a two-stage training paradigm. Stage I injects embodied knowledge into the MLLM through curated digital corpora while maintaining language competence. Stage II trains a policy module through an intent-bridging interface that extracts high-level semantics from the MLLM to guide control, without fine-tuning the MLLM backbone. This process is supported by a self-collected cross-embodiment demonstration suite spanning four robot embodiments and six progressively challenging tasks. Evaluations across digital and physical benchmarks show that a single BLM$_1$ instance outperforms four model families---MLLMs, ELLMs, VLAs, and GMLMs---achieving $\sim\!\textbf{6\%}$ gains in digital tasks and $\sim\!\textbf{3\%}$ in physical tasks.


\begingroup
\newpage
\tableofcontents
\newpage
\endgroup

\section{\textcolor{brainblue}{Introduction}} \label{sec:introduction}

\begin{center}
  \textcolor{brainblue}{\textit{“We must perceive in order to move, but we must also move in order to perceive.”}\\[0.3em]
  \hfill {\small --- James J. Gibson, 1979}}
\end{center}

Recent advances in multimodal large language models (MLLMs)~\cite{MLLM:Survey,MLLM:Qwen2.5-VL,MLLM:GPT-4} have substantially improved joint vision-language reasoning, enabling applications in human-AI interaction, assistive technologies, and embodied intelligence~\cite{MLLM:MLLM-Driving,MLLM:AI-assistant,MLLM:MLLM-Robot}. Leveraging these capabilities, recent works~\cite{VLA:RT-2,VLA:OpenVLA,VLA:Pi0.5} transfer pretrained MLLMs into embodied agents, aiming to bridge digital generalist abilities with physical control.

Two major research lines have emerged. One line of research enhances MLLMs’ embodied reasoning and planning via instruction tuning~\cite{ELLM:Embodiedgpt,ELLM:Cosmos-Reason1}. However, these models cannot directly output robot actions to control the physical embodiments. Another line develops Vision-Language-Action (VLA) models by attaching explicit policy modules, including autoregressive decoders (e.g., RT-2~\cite{VLA:RT-2}, OpenVLA~\cite{VLA:OpenVLA}, FAST~\cite{VLA:FAST}) for discrete action prediction, and diffusion-based heads (e.g., Diffusion Policy~\cite{VLA:DP}, $\pi_{0}$~\cite{VLA:Pi0}, RDT~\cite{VLA:RDT}) for continuous control.
While effective at improving robotic performance, most existing VLA designs~\cite{VLA:OpenVLA, VLA:Pi0, VLA:DiVLA} focus on optimizing robotic performance while \textit{overlooking the retention of MLLM’s generalist reasoning and instruction-following}. To mitigate this, some approaches~\cite{VLA:RT-2, VLA:Pi0.5} adopt a multi-stage training pipeline that combines robot data with multimodal web data: early stages integrate MLLMs with control modules using robot data, interleaved with web data to maintain generalist reasoning, followed by a final robot-only fine-tuning stage that improves task-specific transfer but often impairs instruction-following.
To balance robot control with MLLM’s instruction-following, several studies~\cite{GMLM:ChatVLA-2, GMLM:Robomamba, GMLM:Gemini-Robotics, GMLM:GR00T_N15} freeze MLLMs during the robot-only fine-tuning stage, updating only action modules. 

A central challenge therefore remains: \textit{how to efficiently inject embodied knowledge into MLLMs without degrading their native reasoning and instruction-following, while enabling policy modules to use MLLM-supplied, embodiment-agnostic high-level intent to control diverse physical embodiments.}
To address these challenges, we first formalize three essential capabilities for generalist foundation models: 
\begin{itemize}[leftmargin=1.5em]
\item \textit{Cross-space transfer} maps the knowledge learned by MLLMs in digital domains to physical domains, enabling embodied perception, spatial reasoning, and robotic control.
\item \textit{Cross-task learning} promotes semantic alignment across tasks. For example, embodied question answering reveals object relations, affordances, and causal structure, which enhances planning and execution in long-horizon control tasks.
\item \textit{Cross-embodiment generalization} aligns latent behavior patterns of different embodiments when executing similar tasks, producing a shared latent policy representation. 
\end{itemize}

\begin{figure}[t]
    \centering
    \includegraphics[width=1\textwidth]{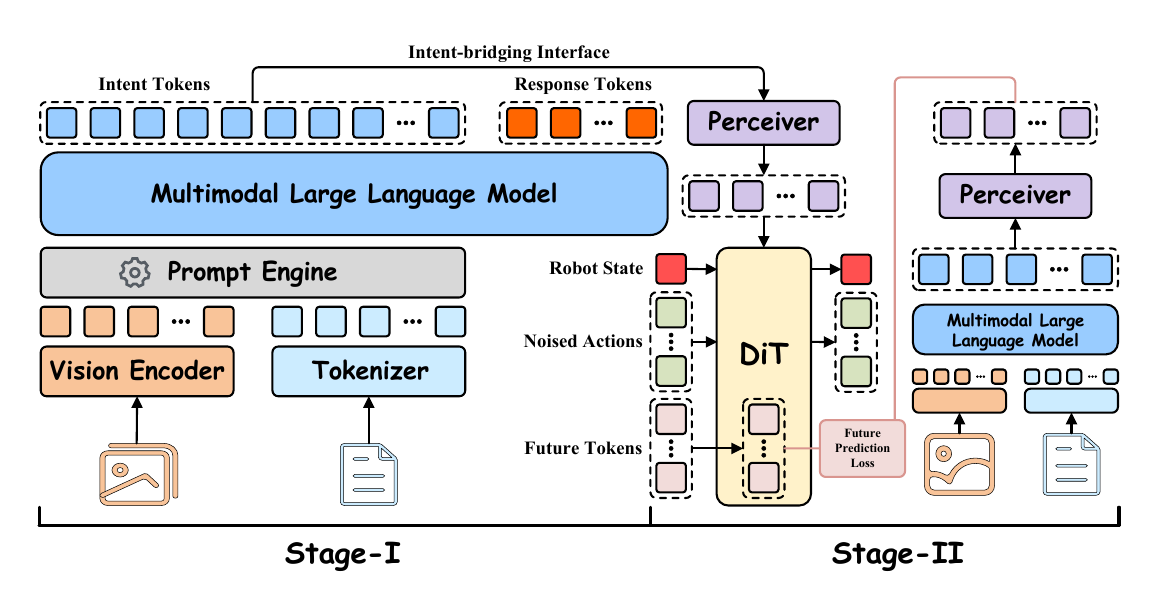}
    \vspace{-5mm}
    \caption{The main framework of BLM$_1$. Multimodal inputs are first encoded and fused by a prompt engine, then passed to the MLLM backbone. BLM$_1$ follows a two-stage training paradigm. In Stage I, the model undergoes supervised fine-tuning on digital-space tasks to acquire embodied knowledge while preserving instruction-following capabilities. Stage II introduces an intent-bridging interface that connects the MLLM to a Diffusion Transformer policy head. This stage is trained using robot states, noisy actions, and a future-prediction loss. The result is \textit{a single unified model} capable of handling both digital and physical tasks, enabling three boundless capabilities: cross-space transfer, cross-task learning, and cross-embodiment generalization.}
    \label{fig:framework}
\end{figure}

Based on the above capabilities, we propose Boundless Large Model with a two-stage training pipeline to efficiently inject embodied knowledge into MLLMs without degrading its instruction-following, then shares a unified diffusion-based policy module to enable efficient cross-embodiment control.
In \textbf{Stage I}, we perform supervised fine-tuning (SFT) for MLLMs on large-scale digital-space understanding and reasoning corpora, endowing MLLMs with physical perception and embodied knowledge while preserving its native instruction-following ability. \textbf{Stage II} freezes MLLMs backbone and trains only a diffusion-based policy head on a self-collected cross-embodiment demonstration suite that spans Franka Emika Panda, xArm-6, xArm-7, and WidowX AI across six increasingly challenging tasks. All demonstrations are generated with the ManiSkill simulator~\cite{ManiSkill3}, which guarantees collision-free trajectories with smooth time parameterization. To the best of our knowledge, this is the first robotic suite that fixes task semantics while systematically varying embodiments to evaluate cross-embodiment generalization. Our main contributions can be summarized as follows:
\begin{itemize}[leftmargin=1.5em]
\item We introduce \textbf{Boundless Large Model (BLM$_1$)}, a multimodal spatial foundation model that unifies three core capabilities: \textit{cross-space transfer, cross-task learning, and cross-embodiment generalization}, referring to Figure~\ref{fig:framework}. BLM$_1$ preserves native instruction-following for question answering and reasoning in the digital space, while supporting seamless cross-embodiment robotic control in the physical space.
\item We propose a systematic two-stage training strategy. Stage I injects embodied knowledge to the MLLM via curated digital corpora. Stage II trains a policy module on cross-embodiment demonstrations via an intent-bridging interface, avoiding MLLM fine-tuning and enabling unified generalization across digital and physical spaces \textit{within a single model}.
\item We evaluate \textit{cross-embodiment generalization} in both digital and physical domains. In the digital space, we assess cross-embodiment question answering and reasoning, and in the physical space, we assess control with a self-collected cross-embodiment demonstration suite on Franka Emika Panda, xArm-6, xArm-7, and WidowX AI across six increasingly challenging tasks.
\item As a \textit{single} model, BLM$_1$ surpasses four major foundation-model families: Multimodal Large Language Models, Embodied Large Language Models, Vision-Language-Action models, and General Multimodal Large Models. It achieves superior performance in digital-space reasoning ($\sim\!\textbf{6\%}$) and physical-space control ($\sim\!\textbf{3\%}$) without model switching.
\end{itemize}

\section{\textcolor{brainblue}{Related Work}} \label{sec:related_work}
\subsection{\textcolor{brainblue}{Multimodal Large Language Models}}
The ascendancy of large language models (LLMs)~\cite{LLM:DeepSeek-R1,LLM:LLaMA,LLM:Mixtral,LLM:Phi-4} has provided a powerful foundation for multimodal large language models (MLLMs)~\cite{MLLM:Qwen2.5-VL,MLLM:LLaVA,MLLM:BLIP-2,MLLM:GPT-4}. 
Beyond contrastive vision-text foundation models~\cite{VFM:CLIP,VFM:ALIGN,VFM:EVA} that specialize in semantic alignment but cannot generate content, MLLMs couple vision foundation models~\cite{VFM:ViT,VFM:MAE,VFM:SigLIP} with a lightweight projector and a language decoder to support end-to-end generation. In practice, they are first pretrained on interleaved image-text corpora to build cross-modal grounding, then aligned by multimodal instruction tuning, and finally refined via preference optimization such as RLHF~\cite{RLHF1} or DPO~\cite{DPO}. This recipe yields models that support a wide range of digital tasks, such as visual question answering, image captioning, and general-purpose multimodal reasoning~\cite{Datasets:GQA,Datasets:MMBench,Datasets:MMMU,Datasets:TextVQA}.

Despite their remarkable progress in multimodal understanding, existing MLLMs~\cite{MLLM:Flamingo,MLLM:Gemini,MLLM:InternVL} are primarily limited to digital domains, lacking the physical perception and spatial reasoning required for real-world tasks. This hinders their applicability in embodied scenarios, where agents must reason about 3D environments, perceive affordances, and generate actionable plans.

\subsection{\textcolor{brainblue}{Embodied Large Language Models}}
To address the physical-world limitations of MLLMs, embodied large language models (ELLMs)~\cite{ELLM:Embodiedgpt,ELLM:Robopoint,ELLM:Palm-e,ELLM:3D-LLM} augment MLLMs with physical perception and embodied reasoning, enabling the transfer of general-purpose reasoning to physical tasks for robotic agents~\cite{Inner_monologue,CodeAP,SayCan}. 
Recent works on ELLMs follow two main routes. The first~\cite{ELLM:Point-LLM,ELLM:3D-LLM,ELLM:LEO} strengthens spatial perception by incorporating explicit 3D inputs into foundational MLLMs, such as point-cloud, depth, or voxel representations. These representations provide richer geometric priors that help the model ground instructions in physical space. The second~\cite{ELLM:Magma,ELLM:Cosmos-Reason1,ELLM:RoboBrain,ELLM:RoboBrain2,ELLM:VeBrain} preserves the input modality and improves embodied planning and reasoning by training on large embodied corpora, often with chain-of-thought guidance~\cite{CoT,CoT1}. Through instruction tuning, this route encourages MLLMs to produce more structured and transparent embodied reasoning and stepwise plans.

However, ELLMs focus on high-level task planning rather than grounded action generation and low-level control. As a result, they often operate in an open-loop mode with weak coupling between perception and action, making it hard to convert abstract plans into reliable continuous control signals. To deploy MLLMs in the physical world, some works typically either fine-tune~\cite{ELLM:Magma,ELLM:Palm-e} ELLMs on robotic demonstrations or attach separate action-generation modules~\cite{ELLM:VeBrain,ELLM:Robopoint}. The former degrades instruction-following, while external modules create a fragmented pipeline that undermines end-to-end closed-loop optimization.

\subsection{\textcolor{brainblue}{Vision-Language-Action Models}}
Vision-language-action models (VLAs)~\cite{VLA:OpenVLA,VLA:Pi0,VLA:GR-2,VLA:Octo} integrate an MLLM backbone with a policy head to map visual observations and natural language instructions to executable control. Policy heads in practice adopt two main paradigms: autoregressive next-token decoding over discrete action tokens, and diffusion-based policies that generate continuous trajectories.
For the autoregressive paradigm~\cite{VLA:RT-2,VLA:OpenVLA,VLA:RT-1}, a subset of the tokenizer vocabulary is overwritten with action tokens, and the backbone decodes discrete control tokens step by step. This discretization imposes limits on high-frequency and dexterous operations, and token-by-token decoding can accumulate long-horizon errors.
In contrast, the diffusion paradigm~\cite{VLA:Pi0.5,VLA:RDT,VLA:dita,VLA:DiVLA} employs a denoising policy module that generates continuous action sequences, typically yielding smoother control signals at higher inference frequencies. However, the additional module training introduces incompatibility between the data distributions of web-scale multimodal corpora and robotic demonstrations, making joint training inefficient.
To improve the generalization of policy heads following the above paradigms and built on general-purpose MLLMs, several studies~\cite{VLA:HPT,VLA:UniACT,VLA:GR00T_N1} explore shared semantics across agents to enable transferable control policies. They often employ embodiment-specific encoders and decoders alongside a shared policy backbone to support adaptation while preserving generalization.

Despite these advances, existing paradigms often undermine the instruction-following and generalist reasoning abilities of MLLMs when adapted to robot-specific tasks. Autoregressive methods often replace a subset of language tokens directly with action tokens, which restricts the expressiveness of the policy and makes it difficult to achieve accurate control across diverse embodiments. Diffusion-based methods rely on separately trained policy modules that are loosely coupled with the MLLM backbone, making it difficult to preserve the model’s original language capabilities during transfer to robot-specific tasks.
Therefore, existing approaches lack a unified foundation model that enables both instruction execution and cross-embodiment policy transfer, limiting their scalability to heterogeneous agents.

\subsection{\textcolor{brainblue}{General Multimodal Large Models}}
General multimodal large models (GMLMs)~\cite{GMLM:ChatVLA,GMLM:Gemini-Robotics,GMLM:Robomamba} leverage a unified architecture that integrates vision, language, and proprioception, enabling the processing of multimodal inputs for both high-level reasoning and low-level control within a single model. Unlike VLAs that center on low-level control and ELLMs that stress abstract embodied reasoning, GMLMs unify language-grounded reasoning with embodied action, delivering both interpretable plans and executable control.
Mainstream GMLM methodologies can be categorized into two principal paradigms.
The first involves architecture-centric strategies~\cite{GMLM:ChatVLA,GMLM:ChatVLA-2}, which fuse MLLMs~\cite{MLLM:BLIP-2,MLLM:Qwen2.5-VL,MLLM:Flamingo,MLLM:PaLI} with robotic systems through multi-stage training and mixture-of-experts mechanisms~\cite{MoE1,MoE2}.
The second includes generalist embodied frameworks~\cite{GMLM:GR00T_N15,GMLM:Robomamba,GMLM:Gemini-Robotics} that freeze MLLMs to preserve instruction-following capabilities while enabling real-world robotic control.

Despite these advances, current GMLMs still face significant challenges. One major issue lies in cross-space embodied knowledge transfer, generalizing reasoning and control across diverse embodiments, and maintaining consistent performance across heterogeneous tasks. In addition, existing models lack mechanisms for rapidly injecting embodied knowledge and seamlessly bridging high-level intent with low-level policy execution, which hinders unified control across different robotic platforms.

\section{\textcolor{brainblue}{BLM$_1$: Boundless Large Model}} \label{BLM-0}
\paragraph{Overview.} In this section, we present the overall design of \textbf{BLM$_1$} referring to Figure~\ref{fig:framework}, which integrates MLLMs with physical perception, spatial reasoning, and embodied control through a two-stage training paradigm. In \textbf{Stage I}, we perform supervised fine-tuning on corpora of digital-space tasks (e.g., spatial reasoning, affordance prediction). \textbf{Stage II} further trains BLM$_1$ for cross-embodiment physical control, while preserving the instruction-following established in Stage I. After training, BLM$_1$ unifies digital-space reasoning and physical-space control within a single model, and seamlessly generalizes across embodiments.

\subsection{\textcolor{brainblue}{Model Architecture}} 
\paragraph{Backbone.}
We use the open-source Qwen2.5-VL-7B-Instruct~\cite{MLLM:Qwen2.5-VL} model as the MLLM backbone for digital-space tasks. Given sampled video frames\footnote{The detailed sampling strategy refers to Section~\ref{sec:frame_sampling}.} $x\!\in\!\mathbb{R}^{t\times h\times w\times 3}$ and a text prompt $y$, the backbone generates a sequence of response tokens $\varphi=(\varphi_1,\dots,\varphi_N)$ and corresponding multimodal hidden states. We extract the multimodal hidden state $H_k\!\in\!\mathbb{R}^{L\times d}$ from the $k$-th transformer block that summarizes high-level intent for control. 

\paragraph{Intent-bridging interface.}  
To effectively bridge high-level intent with low-level control for generalizable policy execution, we design manipulation prompts for the MLLM backbone based on a multi-view historical observation window. To distill essential control-relevant information from dense intent tokens, we integrate a Perceiver~\cite{perceiver} module to compress the high-level intent representation $H_k$ into $\tilde{H_k}$ with fixed $K$ tokens. This token compression prioritizes guidance semantics while also reducing computational overhead, enabling faster processing and supporting high-frequency closed-loop control.

\paragraph{Robot policy.}
For continuous robotic control, we use a vanilla diffusion transformer (DiT)~\cite{VLA:DP,VLA:GR00T_N1} conditioned on the compressed $\tilde{H_k}$ from the MLLM. The current proprioceptive state $q_t$ is encoded through a lightweight state encoder $f_s$. The action chunk over a prediction horizon of $h$ is defined as $A_t = [\mathbf{a}_t, \mathbf{a}_{t+1}, \dots, \mathbf{a}_{t+h-1}]$. $A_t$ is perturbed with noise and subsequently embedded by a lightweight action encoder $f_a$. DiT then performs denoising via interleaved self-attention across the state and action embeddings, together with cross-attention on $\tilde{H_k}$. A lightweight action decoder finally reconstructs the denoised action chunk for execution. The overall module contains approximately 0.76B parameters. To facilitate cross-embodiment generalization, DiT parameters are shared across different embodiments, whereas the state encoders and action encoders/decoders remain embodiment-specific.

\subsection{\textcolor{brainblue}{Training Objective}}
\paragraph{SFT Objective.}
The MLLM backbone with trainable parameters $\theta_{mllm}$ is adapted via standard supervised fine-tuning, i.e., next-token cross-entropy loss:
\begin{equation}
\mathcal{L}_{\mathrm{SFT}}
= -\,\mathbb{E}_{(x,y,\varphi)\sim\mathcal{D}}
\left[
\frac{1}{\sum_{i=1}^{N} m_i}
\sum_{i=1}^{N} m_i \,\log p_{\theta}\!\left(\varphi_i \mid \varphi_{<i},\,x,\,y\right)
\right],
\label{eq:sft}
\end{equation}
\noindent where $m_i \in \{0,1\}$ denotes the supervision mask, $N$ is the token number, $\varphi_{<i}$ represents all previous tokens, $x$ and $y$ denote the sampled video frames and text prompt, $\mathcal{D}$ is the dataset of training samples, and $p_\theta(\cdot)$ is the model's conditional probability.

\paragraph{Flow-matching Objective.}
We train with \textit{flow-matching}~\cite{FlowMatching} under a rectified-flow path. Actions are standardized by dataset statistics $\mu$ and $\sigma$ as
\begin{equation}
\bar{\mathbf{a}} = (\mathbf{a} - \mu) \oslash \sigma,
\end{equation}
where $\mu$ and $\sigma$ denotes the mean and the standard deviation, and $\bar{\mathbf{a}}$ denotes standardized action. We sample $z \sim \mathcal{N}(0, I)$ and $\tau \sim \mathrm{Unif}(0,1)$, and define:
\begin{equation}
\mathbf{x}_\tau = (1 - \tau) \, z + \tau \, \bar{\mathbf{a}}, \qquad
\frac{d\mathbf{x}_\tau}{d\tau} = \bar{\mathbf{a}} - z \;\triangleq\; \mathbf{v}^{\star},
\end{equation}
where $\mathbf{x}_\tau$ evolves from $z$ to $\bar{\mathbf{a}}$, and $\mathbf{v}^{\star}$ is the target velocity.
The DiT parameterizes the vector field $\mathbf{v}_\phi(\mathbf{x}_\tau, \tau | \tilde{H_k})$, and the loss is:
\begin{equation}
\mathcal{L}_{\mathrm{FM}}(\phi) = \mathbb{E}_{(\mathbf{a}, \tilde{H_k}),\,z,\,\tau} \left[ \big\| \mathbf{v}_\phi(\mathbf{x}_\tau, \tau | \tilde{H_k}) - (\bar{\mathbf{a}} - z) \big\|_2^2 \right].
\label{eq:fm}
\end{equation}
At inference, we integrate ${d\mathbf{x}_\tau}/{d\tau} = \mathbf{v}_\phi(\mathbf{x}_\tau, \tau | \tilde{H_k})$ from $\tau = 0$ to $1$ and de-standardize $\hat{\mathbf{a}} = \sigma \odot \mathbf{x}_1 + \mu$.

\paragraph{Future Prediction Objective.}
Inspired by FLARE~\cite{flare}, we introduce a future prediction loss in the later phase of Stage II as an auxiliary objective. Specifically, we extend the DiT input sequence by adding $K$ learnable future tokens \( \mathbf{F} \) and extract their latent representations from an intermediate $l$-th layer of DiT parameterized by $f(\cdot)$. These latent representations are aligned with the $K$ compressed tokens from the future observation $o_{t+H}$ and instruction prompt $y$ via a frozen MLLM backbone, assisted by a lightweight Perceiver module $g(\cdot)$ in BLM$_1$. The future prediction loss is formally defined as:
\begin{equation}
\mathcal{L}_{\mathrm{FP}} = - \mathbb{E}_{\tau}\Big[ \cos \big( f(q_t, A^{\tau}_t, \mathbf{F} | \tilde{H_k}), \, g(o_{t+H}, y) \big) \Big].
\label{eq:fp}
\end{equation}

\paragraph{Overall Objective.}
In Stage I, $\mathcal{L}_{\mathrm{SFT}}$ is used to update LLM and freeze vision encoder and projector in MLLMs. In Stage II, we freeze the MLLMs. During the early phase of Stage II, we use $\mathcal{L}_{\mathrm{FM}}$ to update the Perceiver, DiT, and the projectors (state encoders, action encoders/decoders). In the later phase, we apply the objective $\mathcal{L}_{\mathrm{FM}} + \lambda\,\mathcal{L}_{\mathrm{FP}}$ to further update these components, enabling the model to leverage both flow-matching and future prediction losses for improved performance.

\subsection{\textcolor{brainblue}{Training Recipe}}

\paragraph{Data Weighted Sampling.} 
To balance contributions from different training datasets, we adopt a weighted sampling strategy that enforces equal dataset-level probability while keeping uniform sampling within each dataset. Let \(D_i\) be the \(i\)-th dataset with \(|D_i|\) samples. We assign per-sample weights so that the sampling probability becomes
\begin{equation}
p(x) = \frac{1}{C \cdot |D_i|}, \quad x \in D_i,
\label{eq:weighted_sample_uniform}
\end{equation}
where \(C\) is the number of datasets. Under this scheme, each dataset has equal aggregate sampling probability \(1/C\), and samples are drawn uniformly within their dataset, preventing larger datasets from dominating training updates.

\paragraph{Stage I: Supervised Fine-tuning on MLLMs.} 
We fine-tune the pretrained MLLM backbone on digital multimodal corpora formatted as question–answer pairs. Rather than using action supervision directly, this stage follows the ELLMs paradigm: multimodal QA pairs that span embodied perception and reasoning are used to inject embodied knowledge into the MLLMs, while preserving their general reasoning and instruction-following abilities.

\paragraph{Stage II: Cross-Embodiment Learning.}
We freeze the MLLM backbone from Stage I and apply the flow-matching loss in Equation~\ref{eq:fm} across diverse robotic embodiments. Each embodiment uses its own data for optimizing the embodiment-specific state encoders and action encoders/decoders, while the shared DiT module is optimized across all embodiments for unified action generation. In the early phase of Stage II, we train the Perceiver solely to compress MLLM outputs for DiT conditioning, ensuring stable representations. In the later phase, the Perceiver remains trainable for MLLM features but is frozen when encoding future observations, preventing gradient leakage. To further improve future prediction, the loss in Equation~\ref{eq:fp} is introduced only in the later phase, enhancing the model's ability to anticipate future states and generalize across embodiments.

\input{Tables/A_summary_of_datasets_used_for_BLM-1_stage_1}
\input{Tables/A_summary_of_benchmarks_used_for_BLM-1}

\section{\textcolor{brainblue}{Experiments}}

\subsection{\textcolor{brainblue}{Digital Dataset and Benchmark}}

\paragraph{Digital Dataset.}
To inject embodied knowledge into BLM$_1$, we perform supervised fine-tuning on a collection of multimodal embodied corpora, including RoboVQA~\cite{robovqa}, AgiBot~\cite{agibot}, HoloAssist~\cite{holoassist}, BridgeData V2~\cite{bridge}, EgoPlan~\cite{egoplan}, and ShareRobot~\cite{ELLM:RoboBrain}. The first four corpora are categorized into \textit{understanding} and \textit{reasoning} subsets, while EgoPlan and ShareRobot contribute additional complex planning scenarios and multimodal embodied question answering tasks. Basic statistics are presented in Table~\ref{table:digital_statistics_train}, and representative examples from each dataset are illustrated in Figures~\ref{fig:RoboVQA_example}-\ref{fig:ShareRobot_example} in the Appendix~\ref{examples_digital_dataset}.

\begin{itemize}[leftmargin=15pt]
\item{\textbf{RoboVQA}}~\cite{robovqa} is a large-scale and highly diverse dataset for robotic visual question answering (VQA), containing 29,520 unique instructions and 829,502 video-text pairs. Each sample consists of task-execution videos recorded from agents (robots and humans), accompanied by instructions and question-answer pairs. RoboVQA organizes questions into six categories: planning, task completion verification, affordance discrimination, affordance generation, past event description, and future event prediction, thereby enabling systematic evaluation of models' reasoning and perceptual capabilities across multiple dimensions.

\item{\textbf{AgiBot}}~\cite{agibot} is a large-scale platform comprising over 1 million trajectories for 217 tasks across five deployment scenarios. Built on the AgiBot G1 hardware platform, it yields AgiBot-World, an open-source robotic manipulation dataset collected by more than 100 homogeneous robots and covering a broad spectrum of real-world tasks. The dataset contains 1,001,552 trajectories totaling 2,976.4 hours, spanning 217 concrete tasks, 87 skills, and 106 scenes.

\item{\textbf{HoloAssist}}~\cite{holoassist} is a large-scale egocentric human-robot interaction dataset that captures the process of two-person collaboration in performing physical manipulation tasks. The dataset spans a total of 166 hours, collected from 350 distinct instructor-executor pairs. It focuses on multi-step, goal-directed manipulation tasks involving 16 categories of objects (e.g., small electronic devices, office appliances, IKEA furniture, laboratory equipment, and industrial equipment), around which 20 physical tasks are designed.

\item{\textbf{BridgeData V2}}~\cite{bridge} covers a broad spectrum of robotic manipulation behaviors. Collected on publicly available, low-cost robotic platforms, it comprises 60,096 trajectories across 24 environments and 100+ object categories, including 50,365 expert demonstrations and 9,731 scripted samples. The dataset spans 13 skill families with varying complexity—emphasizing fundamental skills (e.g., grasp-and-place, push/pull, object reorientation) while also including more complex behaviors (e.g., door opening/closing, drawer opening/closing, surface wiping, cloth folding, block stacking). Overall, the corpus consists of 50,365 expert demonstrations and 9,731 scripted trajectories. 

\item{\textbf{EgoPlan}} combines all multiple-choice questions from EgoPlan-Bench and EgoPlan-IT. EgoPlan-Bench~\cite{egoplan} is an MLLM benchmark designed to evaluate human-level planning capabilities in diverse everyday scenarios from an egocentric perspective, containing 4,939 human-verified multiple-choice questions spanning 3,296 task goals. EgoPlan-IT~\cite{egoplan} is an instruction-tuning dataset with 50K question-answer pairs for human-level planning, constructed from the EPIC-Kitchens~\cite{EPIC-Kitchens} video corpus. Together, these resources enable systematic evaluation of planning and decision-making abilities in real-world settings.

\item{\textbf{ShareRobot}}~\cite{ELLM:RoboBrain} is a high-quality, multi-source heterogeneous dataset providing annotations for task planning, object affordance, and end-effector trajectory. The dataset integrates 23 original datasets from Open X-Embodiment~\cite{Datasets:OXE}, encompassing demonstrations from 102 distinct scenarios (bedrooms, laboratories, kitchens, offices) across 12 robotic embodiments. Statistical analysis reveals 132 categories of atomic actions (e.g., "pick," "move," "reach," "lift"). The dataset provides 1,027,990 planning question-answer pairs, 6,522 affordance annotations, and 6,870 trajectory prediction samples for model training purposes.
\end{itemize}

\begin{figure}[!t]
    \centering
    \includegraphics[width=0.89\textwidth]{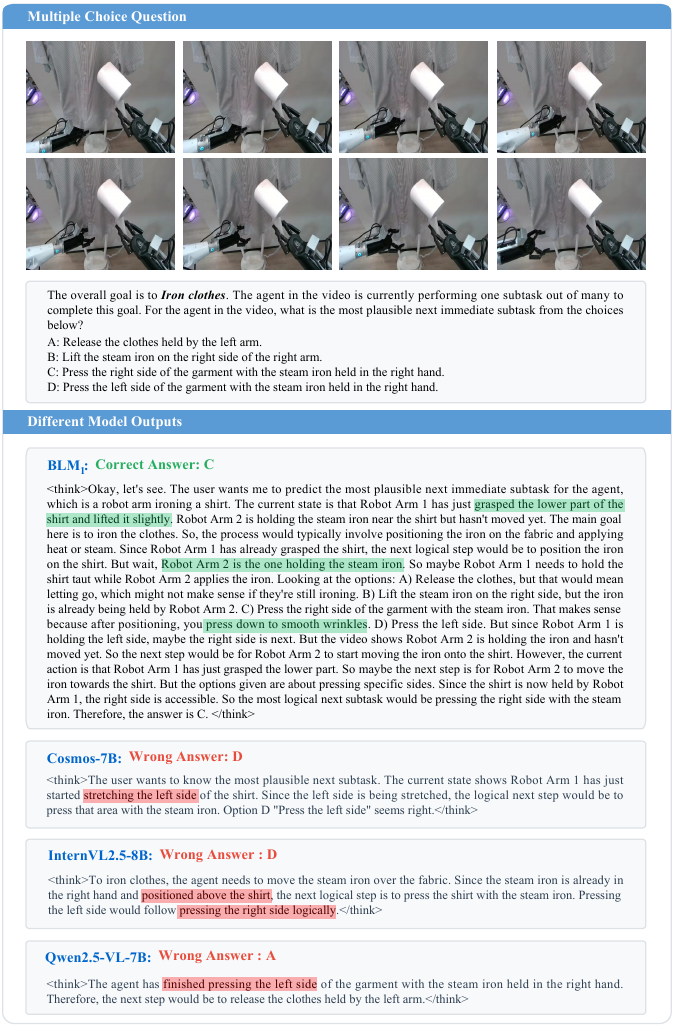}
    \caption{\textbf{Example of results comparison in multiple-choice questions.}}
    \label{fig:benchmark_multiple_choice_QA}
\end{figure}

\begin{figure}[t]
    \centering
    \includegraphics[width=0.87\textwidth]{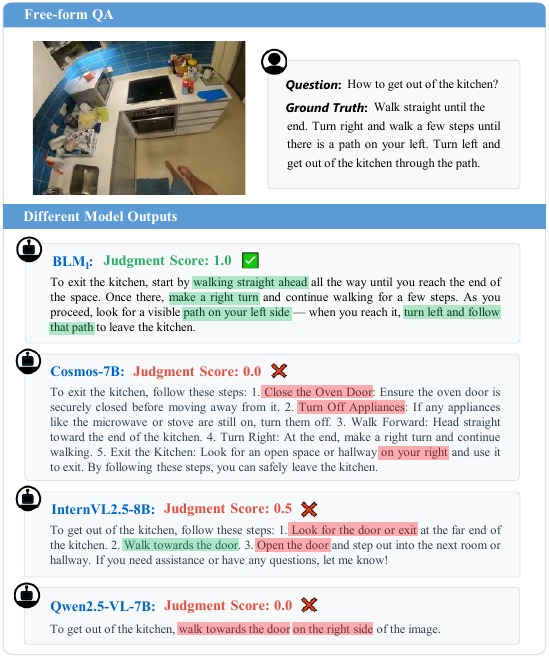}
    \caption{\textbf{Example of results comparison in free-form QA.}}
    \label{fig:benchmark_free_form_QA}
\end{figure}

\paragraph{Digital Benchmark.}
For evaluation on the digital-space tasks, we adopt six commonly used benchmarks: RoboVQA~\cite{robovqa}, AgiBot~\cite{agibot}, HoloAssist~\cite{holoassist}, RoboFail~\cite{robofail}, EgoThink~\cite{egothink}, and ShareRobot~\cite{ELLM:RoboBrain}. Note that BridgeData V2 lacks a test benchmark, while RoboFail provides evaluation data without corresponding training samples. For the remaining datasets, to prevent data leakage and ensure fair evaluation, all original episodes containing test clips are excluded from the training set. The evaluation tasks are structured into two formats: RoboVQA, AgiBot, HoloAssist, and RoboFail use multiple-choice questions, as illustrated in Figure \ref{fig:benchmark_multiple_choice_QA}, whereas EgoThink and ShareRobot employ free-form question-answering, illustrated in Figure \ref{fig:benchmark_free_form_QA}. Basic statistics for these benchmarks are summarized in Table \ref{table:digital_statistics_test}.

\paragraph{Frame Sampling.} 
\label{sec:frame_sampling}
In the processing of video data, a uniform frame sampling strategy is employed, whereby one frame is extracted at 0.5-second intervals. Given that the distributions of video durations vary across datasets (e.g., AgiBot predominantly spans 3$\sim$9 seconds, Bridge 0$\sim$4 seconds, HoloAssist 0$\sim$5 seconds, and RoboVQA is concentrated in the ranges of 2$\sim$3 seconds and 9$\sim$11 seconds), the number of sampled frames is standardized to lie within a range of 4 to 8 per video. Specifically, shorter videos are uniformly sampled at 4 frames, whereas longer videos are uniformly sampled at 8 frames. This design seeks to preserve salient visual information while maintaining balance and comparability in dataset scale.

\subsection{\textcolor{brainblue}{Physical Dataset and Evaluation}}
\paragraph{Robotic Data Collection.}
We employ ManiSkill~\cite{ManiSkill3}, an open-source GPU-accelerated simulation framework for robotic manipulation that enables efficient data collection and supports various learning paradigms. In our setting, we collect demonstrations for six table-top two-finger manipulation tasks using four different robotic embodiments: Franka Emika Panda, xArm-6, xArm-7, and WidowX AI. For each embodiment-task pair, 100 episodes are recorded, including third-person and wrist-view images, proprioceptive states, and actions. The detailed statistics are summarized in Table~\ref{tab:stage2_dataset_summary}, which reports the total number of frames used for each task. Note that the positions and yaw orientations (Z-axis rotations) of the cubes, spheres, and pegs on the table are randomly initialized. The six tasks are defined as follows:

\input{Tables/A_summary_of_datasets_used_for_BLM-1_stage_2}

\begin{figure}[!b]
    \centering
    \includegraphics[width=1\textwidth]{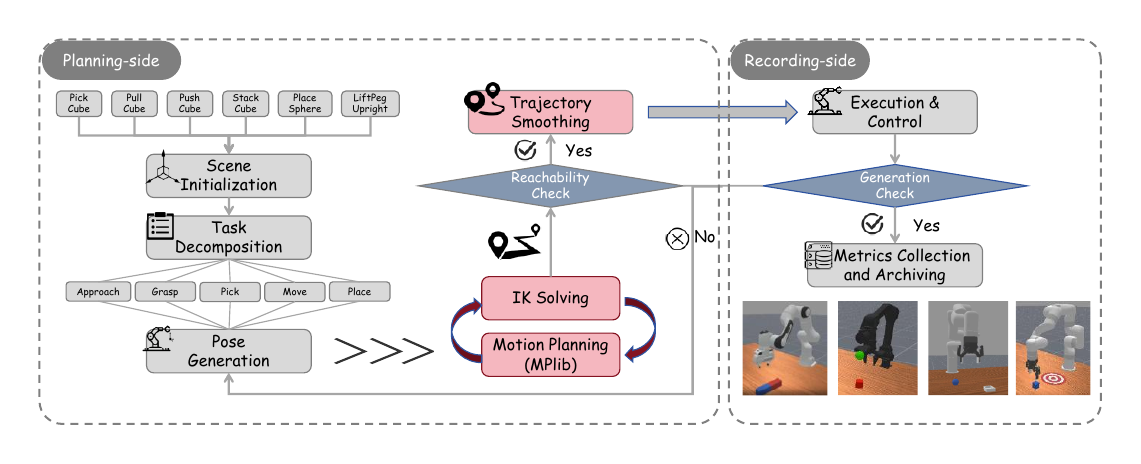}
    \caption{\textbf{Cross-embodiment data collection pipeline for Stage II training.}}
    \label{fig:Stage2_data_collection_pipeline}
\end{figure}

\begin{itemize}[leftmargin=15pt]
\item \textbf{PickCube:} This task requires grasping the red cube from the table with the gripper; success is achieved when the cube is lifted more than 2.5 cm above the table surface.
\item \textbf{PullCube:} This task requires pulling the cube on the table into the designated target region with the gripper; success is achieved when the cube is fully within the target area.
\item \textbf{StackCube: }This task requires grasping the red cube and stacking it on top of the green cube; success is achieved when the red cube is stably placed on the green cube and the gripper is released.
\item \textbf{PushCube:} This task requires pushing the cube on the table into the designated target region with the gripper; success is achieved when the cube is fully within the target area.
\item \textbf{PlaceSphere:} This task requires grasping the blue sphere from the table and placing it into the shallow white container on the table; success is achieved when the sphere is stably placed inside the container.
\item \textbf{LiftPegUpright:} This task requires lifting the peg from the table and making it stand upright; success is achieved when the peg is stably placed vertically on the table.
\end{itemize}

\paragraph{Data Engine.}
Our data collection pipeline is divided into two main stages: planning and recording, as illustrated in Figure~\ref{fig:Stage2_data_collection_pipeline}. In the planning stage, we adopt a unified strategy of high-level skills, key-pose guidance and motion planning execution. Specifically, we first select a sampling task from six predefined robotic manipulation tasks and initialize the corresponding scene. Next, inspired by the concept of keyframes, each task is decomposed into a series of primitive actions (e.g., approaching, grasping, moving), and a target end-effector pose is generated for each primitive. For every target pose, we solve the Inverse Kinematics (IK) to obtain joint configurations and perform feasibility checks, including joint limits, self-collision, and environment collision. If infeasible, resampling or switching to alternative poses is applied. For feasible poses, we conduct path planning and trajectory smoothing before moving to the recording stage. In the recording stage, the robot executes the task either through joint-space trajectories or end-effector pose tracking, while success is continuously monitored. If a failure occurs, the system performs fine adjustments or rollback operations; if successful, the relevant states and trajectories are recorded, completing the data collection process.

\subsection{\textcolor{brainblue}{Evaluation Metrics}} \label{sec:eval_metrics}
\subsubsection{\textcolor{brainblue}{Digital-space Tasks}}
We report two complementary metrics: (i) \textit{exact-match accuracy} for multiple-choice answers, and (ii) an \textit{LLM-as-a-judge} score for free-form answers.

\paragraph{Exact-match accuracy (EM).}
For multiple-choice question answers, the score is computed as follows:
\begin{equation}
\mathbf{S}=\frac{1}{M}\sum_{i=1}^{M}\mathbf{1}\!\left\{\mathrm{Norm}\!\left(a_i\right)=\mathrm{Norm}\!\left(\hat{a}_i\right)\right\},
\end{equation}
where $M$ denotes the total number of samples, $a_i$ denotes the ground-truth answer, and $\hat{a}_i$ denotes the model’s prediction for the $i$-th sample. $\mathbf{1}$ is an indicator function that equals 1 when the prediction matches the ground-truth answer and 0 otherwise. $\mathrm{Norm}(\cdot)$ lowercases and strips punctuation/whitespace with a standard QA normalizer.

\paragraph{LLM-as-a-judge (GPT-based scoring).}
For free-form answers, we use a rubric-guided direct assessment with a held-out judge LLM. 
Given a system prompt $p_s$, a question $q_i$, a human-annotated reference answer $\hat{a}_i$, and a model response $a_i$, the judge outputs a scalar score $s_i$ according to an explicit rubric. The detailed rubric prompts are provided in the Appendix~\ref{sec:scoring_prompts}. The rubric-based scoring procedure is shown as follows:

\begin{itemize}[leftmargin=15pt]
\item \textbf{EgoThink:} This benchmark submits $\left(p_s, q_i, \hat{a}_i, a_i\right)$, together with a unified scoring rubric, to a judge LLM, which returns a fixed discrete score as follows:
\begin{equation}
    \begin{aligned}
        s_i \in \left\{0, 0.5, 1\right\},
    \end{aligned}
\end{equation}
where $s_i=0$ denotes an incorrect answer, $s_i=0.5$ denotes a partially correct answer, and $s_i=1$  denotes a fully correct answer. On this basis, the overall task score can be formalized as:
\begin{equation}
\mathbf{S} = \left( \frac{1}{M} \sum_{i=1}^{M} s_i \right) \times 100.
\end{equation}
\item \textbf{ShareRobot:} Similar to EgoThink, we follow the evaluation protocol of OpenEQA~\cite{openeqa} and submit $\left(p_s, q_i, \hat{a}_i, a_i\right)$, together with a unified scoring rubric, to a judge LLM, which returns a fixed discrete score as follows:
\begin{equation}
    \begin{aligned}
        s_i \in \left\{1,2,3,4,5\right\},
    \end{aligned}
\end{equation}
where $s_i=1$ denotes an incorrect or semantically inconsistent answer, $s_i=5$ denotes a correct answer that is semantically equivalent to the reference, and $s_i\in \left\{2,3,4\right\} $ quantifies increasing levels of partial correctness. On this basis, the overall task score can be formalized as:
\begin{equation}
    \begin{aligned}
        \mathbf{S} = \left( \frac{1}{M} \sum_{i=1}^{M} \frac{s_i - 1}{4} \right) \times 100.
    \end{aligned}
\end{equation}
\end{itemize}

\subsubsection{\textcolor{brainblue}{Physical-space Tasks}}
We use episode-level \textit{success rate} to measure the cross-embodiment performance of policy modules. Each task is evaluated over 50 rollouts, with a task-specific maximum step limit set according to its difficulty. Formally, it is defined as:
\begin{equation}
\mathbf{S} = \frac{1}{E \cdot U} \sum_{e=1}^{E} \sum_{u=1}^{U} \mathrm{sr}_{e,u},
\end{equation}
where $E$ denotes the number of robotic embodiments, $U$ denotes the number of tasks, and the task-level success rate is given by
\begin{equation}
\mathrm{sr}_{e,u} = \frac{\mathrm{N}^{s}_{e,u}}{\mathrm{N}_{e,u}},
\end{equation}
where $\mathrm{sr}_{e,u}$ denotes the proportion of successful rollouts out of $\mathrm{N}_{e,u}=50$ trials for the $u$-th task on the $e$-th embodiment.

\subsection{\textcolor{brainblue}{Quantitative Results}}

\input{Tables/comparison_existing_MLLMs_and_GMLMs_on_spatial_reasoning_benchmarks}

\subsubsection{\textcolor{brainblue}{Quantitative Analysis in Digital Space}}
\paragraph{Overall Comparison on Six Digital-Space Benchmarks.} As shown in Table \ref{table:digital-space benchmarks}, we conduct a systematic evaluation on six digital-space benchmarks, including RoboVQA \cite{robovqa}, AgiBot \cite{agibot}, HoloAssist \cite{holoassist}, RoboFail \cite{robofail}, EgoThink \cite{egothink}, and ShareRobot \cite{ELLM:RoboBrain}. BLM$_1$ achieves an overall average score of \textbf{64.88}, substantially outperforming the strongest closed-source GPT-4o \cite{MLLM:GPT4o} (\textbf{59.86}) and the best-performing open-source Cosmos-7B \cite{ELLM:Cosmos-Reason1} (\textbf{58.55}). Fine-grained subtask results for EgoThink and ShareRobot are reported in Tables \ref{EgoThink} and \ref{ShareRobot}.

Among closed-source MLLMs \cite{MLLM:GPT4o, MLLM:Claude3.5Sonnet}, GPT-4o \cite{MLLM:GPT4o} achieves the best performance on EgoThink (\textbf{72.42}), highlighting its advantage in first-person reasoning. Nevertheless, BLM$_1$ consistently outperforms GPT-4o across all other benchmarks, with significant gains on RoboVQA (\textbf{+20.00}), AgiBot (\textbf{+9.00}), and RoboFail (\textbf{+8.00}). These results suggest that while GPT-4o excels in a narrow reasoning context, BLM$_1$ exhibits broader robustness, driven by systematic advances in high-level planning, affordance reasoning, task verification, action generation, and cross-scenario as well as cross-embodiment generalization.

Within open-source MLLMs \cite{MLLM:LlavaOneVision, MLLM:InternVL, MLLM:Qwen2.5-VL, ELLM:Cosmos-Reason1}, Qwen2.5-VL-7B-Instruct \cite{MLLM:Qwen2.5-VL} and Cosmos-7B \cite{ELLM:Cosmos-Reason1} serve as strong baselines. BLM$_1$ consistently surpasses both models, with average improvements of \textbf{+7.37} over Qwen2.5-VL-7B-Instruct and \textbf{+6.33} over Cosmos-7B. It outperforms them on all six benchmarks, with particularly notable margins on HoloAssist (\textbf{+11.00} over Cosmos-7B) and ShareRobot (\textbf{+18.73} over Qwen2.5-VL-7B-Instruct). These results underscore BLM$_1$'s enhanced embodied multimodal reasoning capabilities, especially in first-person reasoning, planning, and fine-grained action generation.

Compared with embodied large language models (ELLMs) \cite{ELLM:Magma, ELLM:VeBrain}, BLM$_1$ also establishes a clear advantage. Against VeBrain-7B \cite{ELLM:VeBrain}, it delivers consistently strong performance across all tasks, with pronounced gains on RoboVQA (\textbf{+10.00}), HoloAssist (\textbf{+12.00}), and ShareRobot (\textbf{+14.12}). Such gains provide compelling evidence that BLM$_1$ not only improves reasoning and planning but also enhances transferability across tasks, scenarios, and embodiments.

Against general multimodal large models (GMLMs) \cite{GMLM:ChatVLA, ELLM:RoboBrain}, including ChatVLA-2B \cite{GMLM:ChatVLA} and RoboBrain2-7B \cite{ELLM:RoboBrain}, BLM$_1$ consistently achieves superior performance. On average, it achieves an improvement of over \textbf{+11.00}, with particularly pronounced gains on RoboVQA and ShareRobot, further underscoring its effectiveness in embodied multimodal reasoning.

In summary, across all four comparison categories, BLM$_1$ establishes clear superiority in embodied multimodal understanding and reasoning. It surpasses closed-source and open-source MLLMs, achieves substantial improvements over ELLMs, and consistently outperforms GMLMs. These results underscore BLM$_1$'s balanced design, adaptability, and robust generalization across diverse digital-space benchmarks.

\input{Tables/Comparison_with_existing_MLLMs_and_GMLMs_on_EgoThink}

\input{Tables/Comparison_existing_MLLMs_and_GMLMs_on_ShareRobot}

\paragraph{Fine-Grained Analysis on EgoThink and ShareRobot.}
As shown in Table~\ref{EgoThink}, BLM$_1$ achieves an average score of \textbf{59.42} on EgoThink, demonstrating competitive performance in embodied tasks. It excels in object attribute recognition (\textbf{76.00}), surpassing open-source MLLMs with gains of \textbf{+11.00} over both Qwen2.5-VL-7B-Instruct and Cosmos-7B. In planning navigation (\textbf{25.00}), BLM$_1$'s score outperforms them with margins of \textbf{+5.00} over Qwen2.5-VL-7B-Instruct and \textbf{+7.00} over Cosmos-7B, while maintaining balanced capabilities in reasoning tasks like counting (\textbf{56.00}) and situated reasoning (\textbf{59.00}).

Among closed-source MLLMs, GPT-4o achieves the best performance on EgoThink (\textbf{72.42}), highlighting its advantage in first-person reasoning. Nevertheless, BLM$_1$ remains competitive, outperforming Claude-3.5 Sonnet by a substantial margin (\textbf{+10.88}) and demonstrating robust embodied understanding despite trailing GPT-4o. Within open-source MLLMs and ELLMs, BLM$_1$ surpasses peers like Cosmos-7B (\textbf{+2.75}) and Magma-8B (\textbf{+10.59}), underscoring its superior planning efficiency at similar scales. These results position BLM$_1$ as a cost-effective open-source solution for multimodal reasoning in real-world applications.

\input{Tables/comparison_existing_VLAs_on_robot_benchmarks2}
\input{Tables/comparison_existing_VLAs_on_robot_benchmarks}

As shown in Table~\ref{ShareRobot}, BLM$_1$ attains the highest average score of \textbf{65.41} on ShareRobot, excelling in discriminative affordance (positive: \textbf{84.88}) and future prediction (\textbf{73.29}). It leads in planning with context (\textbf{85.73}) and success judgment (positive: \textbf{71.22}), outperforming GPT-4o in categories like generative affordance (\textbf{29.15} vs. \textbf{29.02}) and demonstrating robust task planning (\textbf{60.49}) with remaining steps prediction (\textbf{72.80}).

Among closed-source MLLMs, BLM$_1$ outperforms GPT-4o (\textbf{+4.11}) on ShareRobot, highlighting its advantages in affordance reasoning and predictive planning. Within open-source MLLMs, it surpasses Cosmos-7B (\textbf{+7.73}) and Qwen2.5-VL-7B-Instruct with even larger gains. Compared with ELLMs like VeBrain-7B and general multimodal large models like RoboBrain2-7B, BLM$_1$ achieves consistent superiority, emphasizing its exceptional accuracy and generalization in robot manipulation tasks across comparable parameters.

\subsubsection{\textcolor{brainblue}{Quantitative Analysis in Physical Space}}
As shown in Tables \ref{tab:physical_space} and \ref{tab:physical_space_all}, we conduct a comprehensive evaluation on four robot embodiments (Panda, xArm-6, xArm-7, and WidowX AI) across six progressively challenging tasks, comparing against representative baselines including $\pi_0$~\cite{VLA:Pi0}, HPT~\cite{VLA:HPT}, UniAct~\cite{VLA:UniACT}, Diffusion Policy~\cite{VLA:DP}, GR00T-N1~\cite{VLA:GR00T_N1}, and GR00T-N1.5~\cite{GMLM:GR00T_N15}. BLM$_1$ achieves an average success rate of \textbf{75.83\%}, surpassing all from-scratch policies and even outperforming pre-trained VLAs, highlighting its superior cross-embodiment generalization across diverse tasks.

At the embodiment level, BLM$_1$ consistently outperforms both pre-trained and from-scratch models across all four robotic platforms. For pre-trained baselines, the GR00T series still falls short of BLM$_1$, while $\pi_0$ only achieves \textbf{67.42\%} (or \textbf{66.08\%} for $\pi_0^\dagger$), showing that large-scale semantic priors and flow-matching alone cannot guarantee robust cross-embodiment generalization. Similarly, HPT and UniAct, which aim for embodiment-agnostic generalization, deliver much lower results (e.g., HPT scores \textbf{55.00\%} on Panda and \textbf{48.67\%} on xArm-6), with BLM$_1$ surpassing them by large margins (\textbf{+29.67\%} and \textbf{+31.66\%}, respectively). Compared to from-scratch methods, BLM$_1$ also demonstrates clear advantages. Diffusion Policy requires specialized training to achieve competitive results, but still trails BLM$_1$ in overall average (\textbf{71.67\%} vs. \textbf{75.83\%}). Concretely, DP slightly surpasses BLM$_1$ on xArm-7, but BLM$_1$ shows higher success rates on Panda (\textbf{+10.67\%}), xArm-6, and WidowX AI, while also outperforming GR00T-N1.5 on all platforms (e.g., \textbf{+6.00\%} on xArm-6, \textbf{+4.00\%} on xArm-7). Even on the more challenging WidowX AI, BLM$_1$ achieves \textbf{63.67\%}, higher than all baselines. Overall, BLM$_1$ maintains consistent performance across platforms, showing robust cross-embodiment generalization despite differences in joint count, workspace, and visual perception.

We further conduct a task-level analysis to examine how different models maintain performance across task details and embodiment variations. As shown in Table~\ref{tab:physical_space_all}, BLM$_1$ exhibits strong task robustness and embodiment generalization. On relatively simple tasks such as PickCube, PushCube, and PullCube, it achieves an average success rate of \textbf{83.5\%}, the highest among all compared models, reflecting near-perfect handling of basic action prediction and perception as well as robust execution of simple tasks. On more challenging tasks such as PlaceSphere and LiftPegUpright, BLM$_1$ still maintains an average success rate of \textbf{60.5\%}, ranking best among from-scratch models and outperforming most pre-trained counterparts, with overall performance comparable to GR00T-N1.5. These results highlight BLM$_1$’s consistent stability as task complexity increases.

In summary, BLM$_1$ demonstrates clear advantages at both the embodiment and task levels. It consistently outperforms representative pre-trained and from-scratch models across four distinct embodiments, showing remarkable cross-embodiment generalization. At the same time, it sustains high success rates across tasks of increasing difficulty, highlighting task-level stability. Together, these results establish BLM$_1$ as a unified policy that excels in cross-embodiment generalization and task robustness, ensuring stable and superior performance across diverse settings.

\begin{figure}[t]
    \centering
    \includegraphics[width=0.92\textwidth]{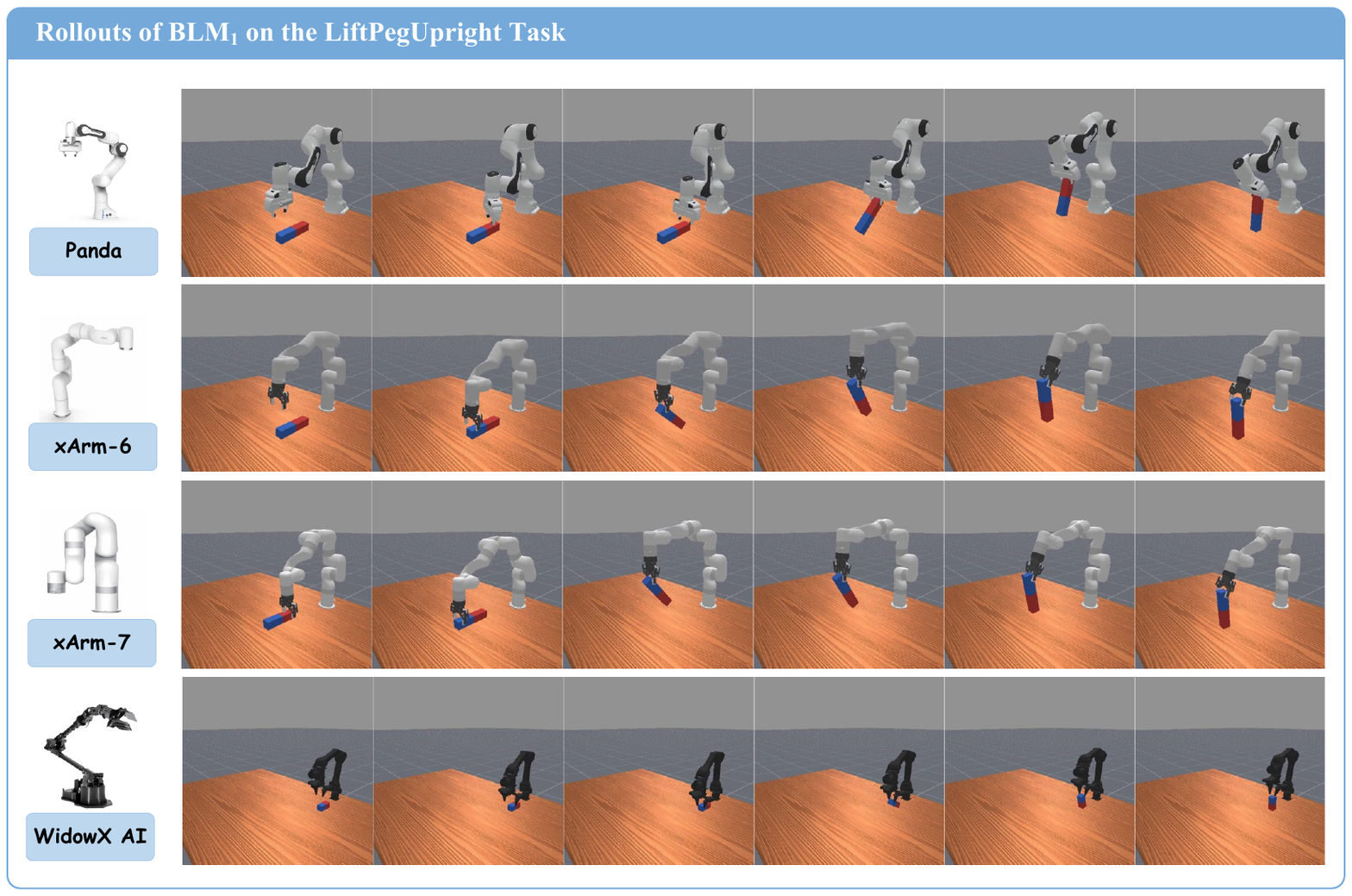}
    \caption{\textbf{Visualization of cross-embodiment rollouts with \textbf{BLM$_1$} on the LiftPegUpright task.}}
    \label{fig:reasoning_sample_robot}
\end{figure}

\subsection{\textcolor{brainblue}{Qualitative Results}}
As illustrated in Figure~\ref{fig:benchmark_multiple_choice_QA}, our BLM$_1$ demonstrates superior reasoning capabilities in multiple-choice questions by accurately identifying the most plausible next subtask in a robotic ironing scenario. While other models, such as Cosmos-7B and InternVL2.5-8B, erroneously select option D (pressing the left side) due to misinterpreting the robot arms' positions and the ongoing action, and Qwen2.5-VL-7B chooses A (releasing the clothes) by prematurely assuming task completion, BLM$_1$ correctly opts for C (pressing the right side). This advantage stems from BLM$_1$'s precise analysis of the visual context, recognizing that Robot Arm 1 has grasped the lower part of the shirt, making the right side accessible for ironing by Robot Arm 2. Consequently, BLM$_1$ avoids common pitfalls like spatial confusion or logical sequencing errors, ensuring alignment with the overall goal of ironing clothes. Such robustness highlights BLM$_1$'s enhanced multimodal understanding and step-by-step reasoning, outperforming baselines in complex task prediction.

In Figure~\ref{fig:benchmark_free_form_QA}, BLM$_1$ excels in free-form question answering by providing detailed, accurate navigation instructions that closely match the ground truth, achieving a judgment score of 1.0. In contrast, InternVL2.5-8B offers vague, generic steps with a score of 0.5, failing to capture specific visual cues like the path layout; Cosmos-7B hallucinates irrelevant actions such as closing the oven door and turning off appliances, resulting in a 0.0 score; and Qwen2.5-VL-7B misdirects to a non-existent door on the right, also scoring 0.0. BLM$_1$'s response meticulously describes walking straight to the end, turning right, then left onto the path, demonstrating superior scene comprehension and avoidance of hallucinations. This precision underscores BLM$_1$'s strength in integrating visual details with contextual reasoning, yielding reliable, step-specific outputs that surpass other models in real-world applicability for tasks like spatial navigation.

\section{\textcolor{brainblue}{Conclusion}}
In this paper, we introduce BLM$_1$, a unified spatial foundation model that bridges digital and physical spaces by integrating embodied reasoning, instruction following, and closed-loop robotic control.
Through a two-stage training pipeline, BLM$_1$ acquires both embodied knowledge and low-level control capabilities, while preserving the native language understanding of its MLLM backbone.
Experimental results show that BLM$_1$ improves performance by 6\% on digital-space benchmarks and 3\% on real-world robotic tasks, while also generalizing across diverse tasks and embodiments.
By jointly enabling cross-space transfer, cross-task adaptation, and cross-embodiment control within a single model, BLM$_1$ presents a scalable path toward general-purpose embodied intelligence.

\clearpage

\bibliographystyle{unsrt}  
\bibliography{references} 

\clearpage

\appendix

\section{\textcolor{brainblue}{Dataset Examples}}

\subsection{\textcolor{brainblue}{Examples for Digital Dataset}}
\label{examples_digital_dataset}

\begin{figure}[!b]
    \centering
    \includegraphics[width=1\textwidth]{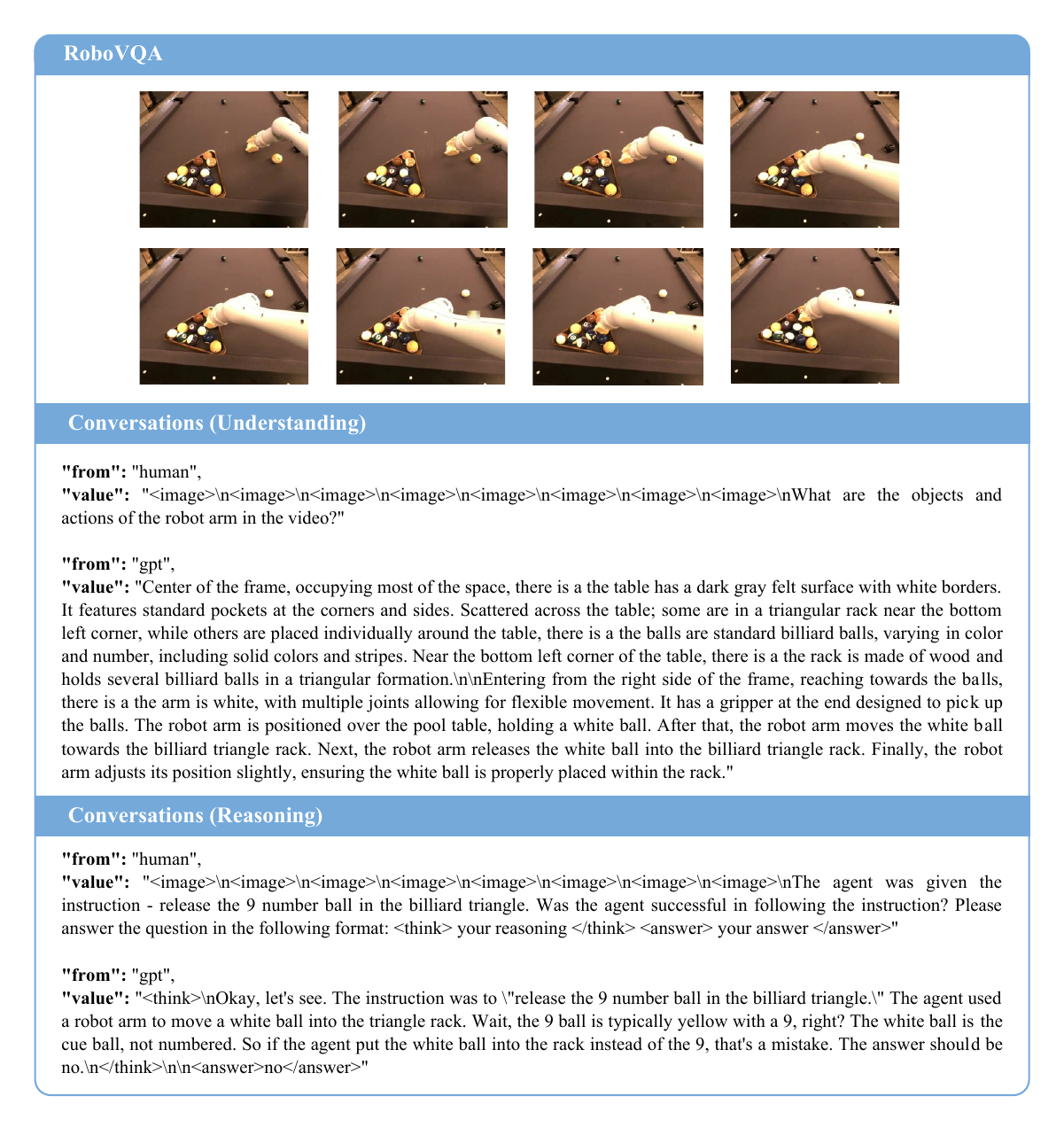}
    \caption{\textbf{Visualization of training example from RoboVQA.}}
    \label{fig:RoboVQA_example}
\end{figure}

\begin{figure}[!b]
    \centering
    \includegraphics[width=1\textwidth]{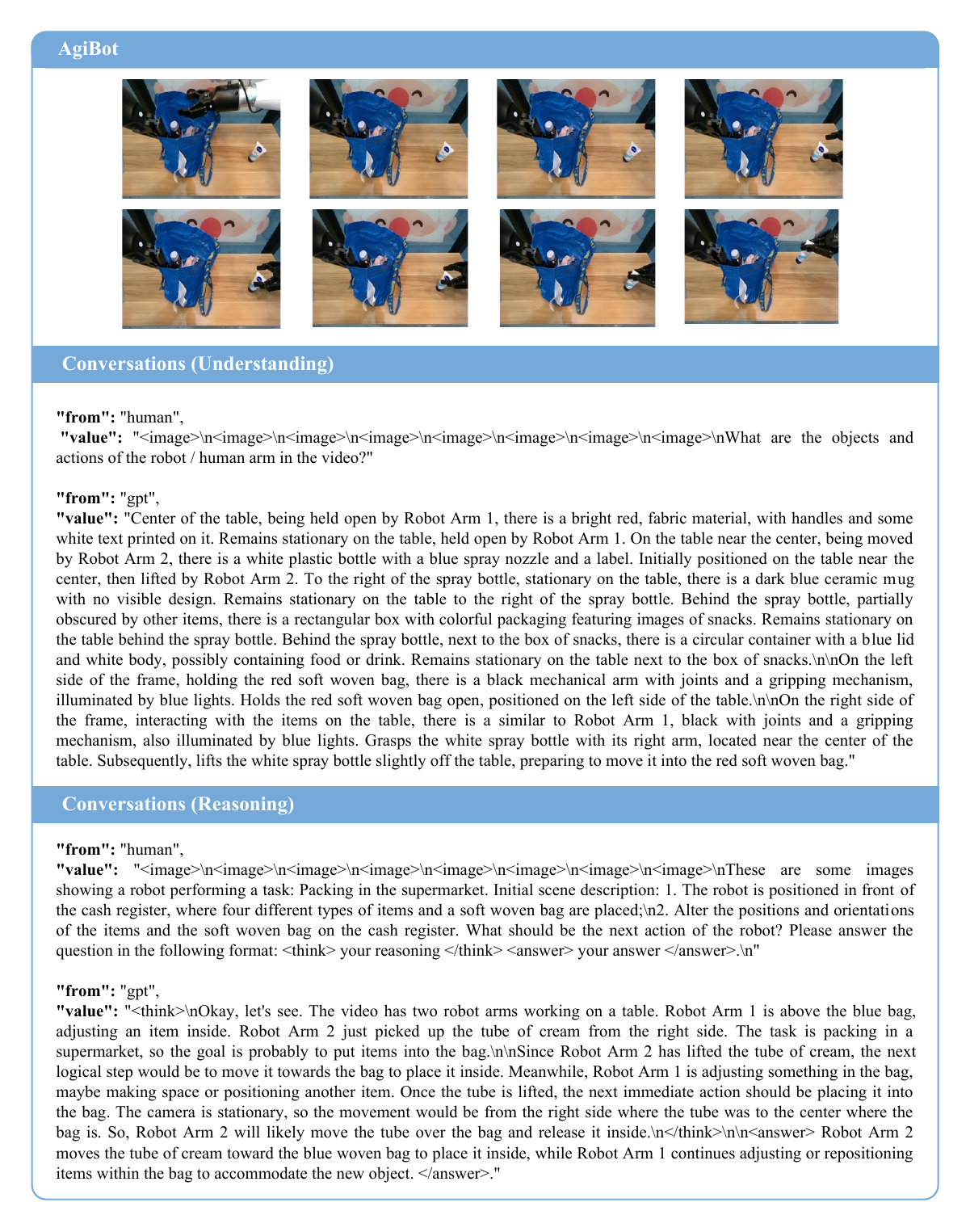}
    \caption{\textbf{Visualization of training example from AgiBot.}}
    \label{fig:AgiBot_example}
\end{figure}

\begin{figure}[!b]
    \centering
    \includegraphics[width=1\textwidth]{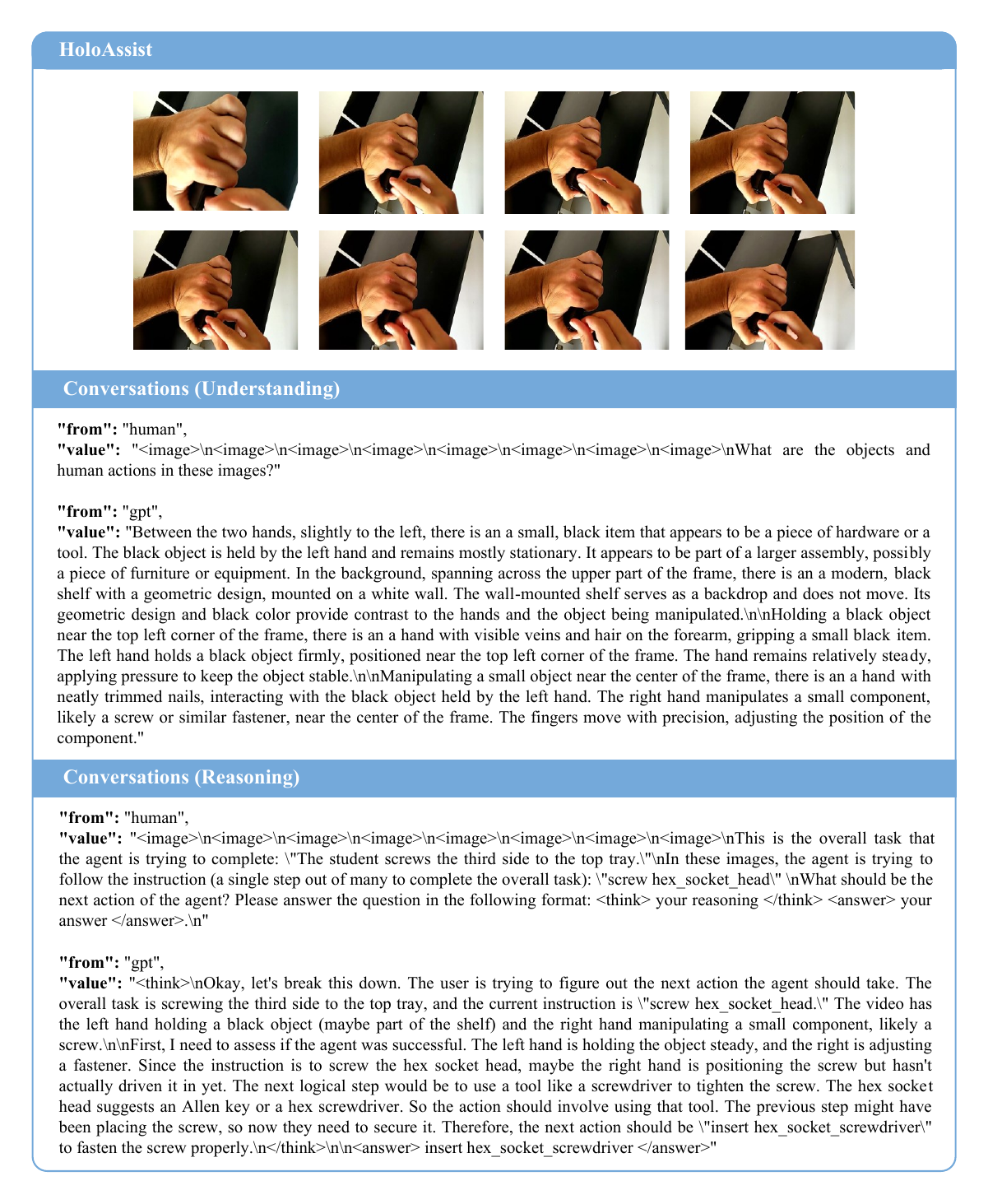}
    \caption{\textbf{Visualization of training example from HoloAssist.}}
    \label{fig:HoloAssist_example}
\end{figure}

\begin{figure}[!b]
    \centering
    \includegraphics[width=1\textwidth]{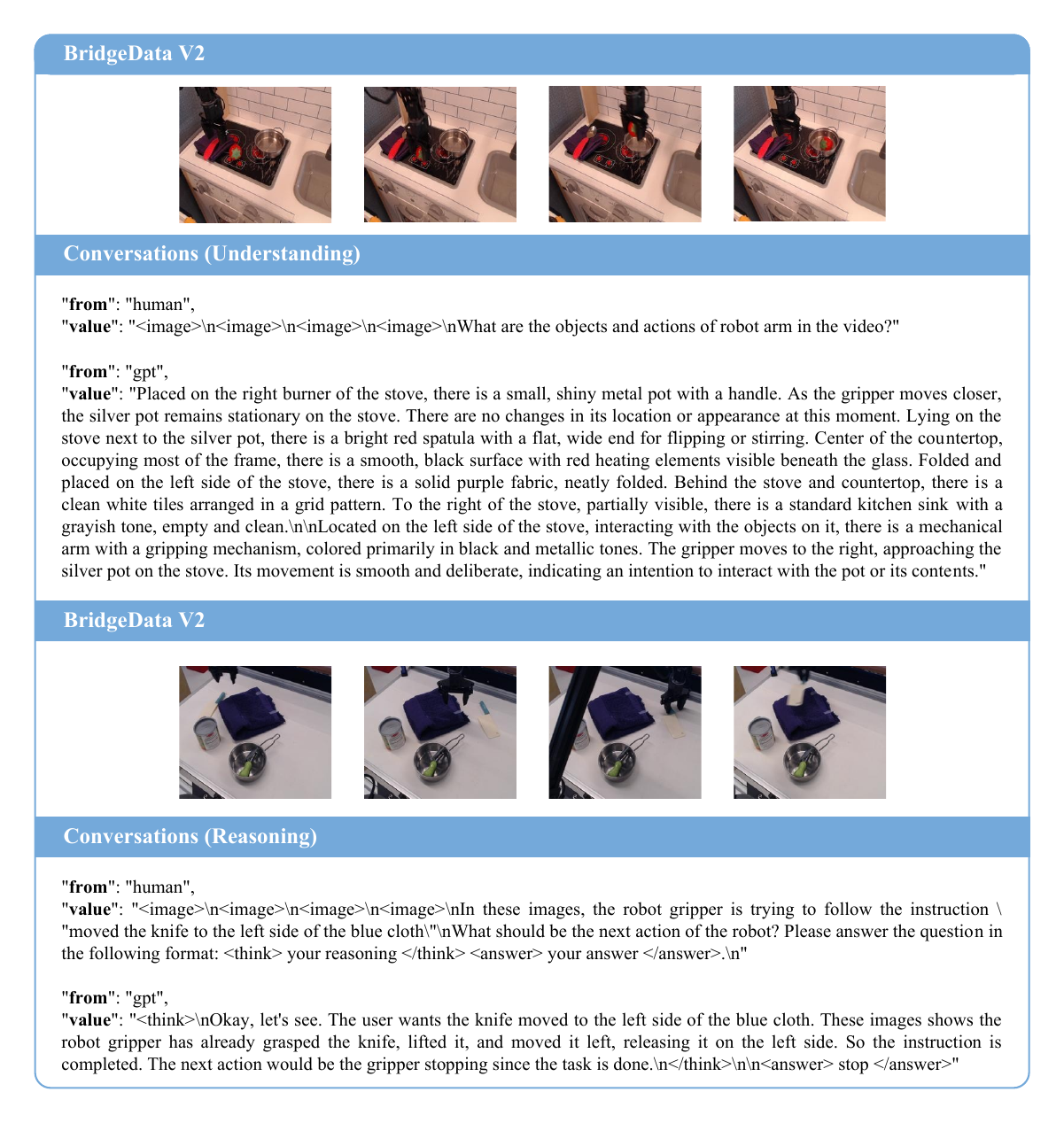}
    \caption{\textbf{Visualization of training example from BridgeData V2.}}
    \label{fig:BridgeData_example}
\end{figure}

\begin{figure}[!b]
    \centering
    \includegraphics[width=1\textwidth]{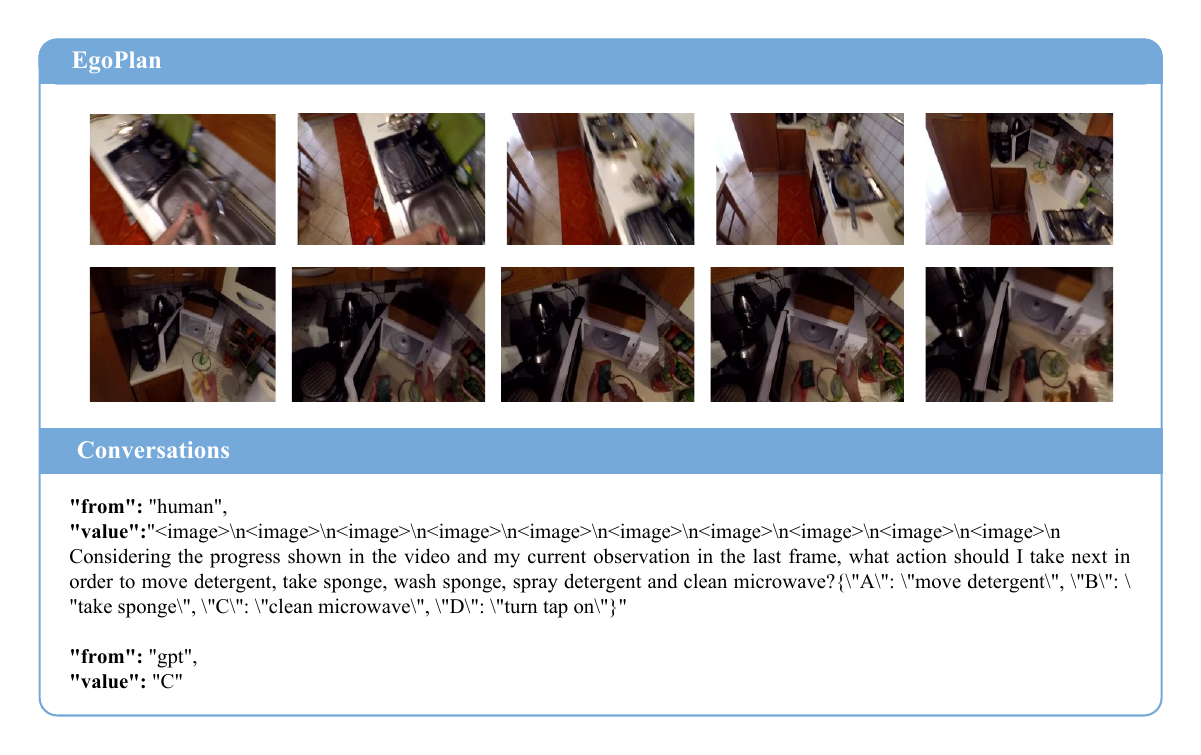}
    \caption{\textbf{Visualization of training example from EgoPlan.}}
    \label{fig:EgoPlan_example}
\end{figure}

\begin{figure}[!b]
    \centering
    \includegraphics[width=1\textwidth]{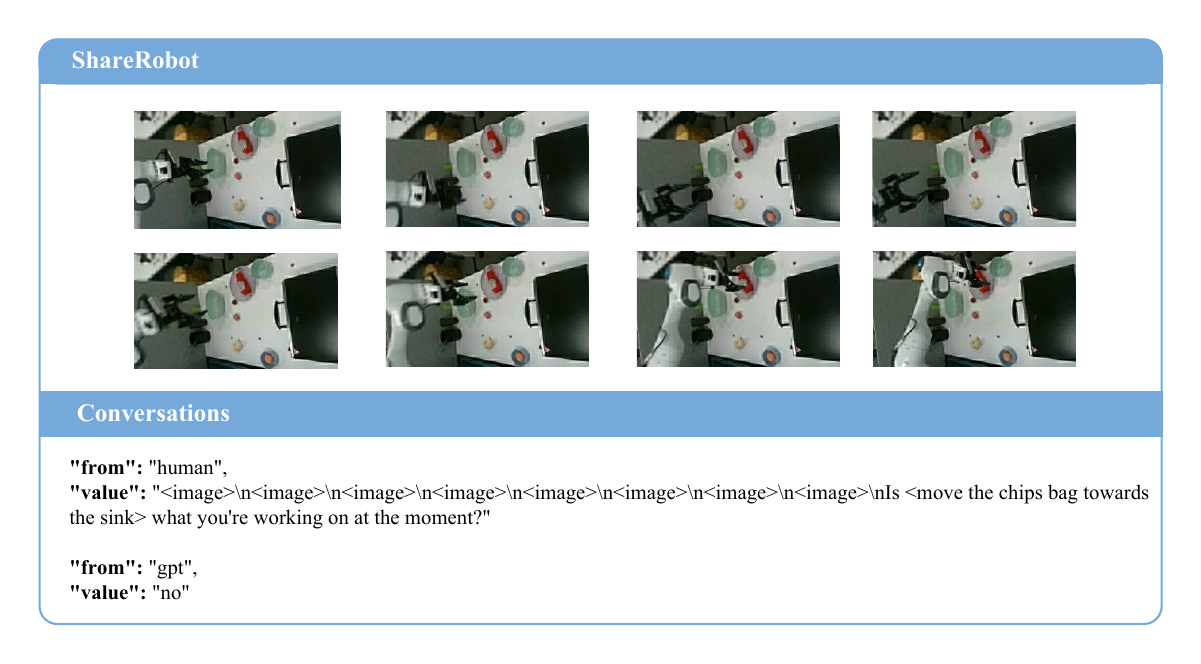}
    \caption{\textbf{Visualization of training example from ShareRobot.}}
    \label{fig:ShareRobot_example}
\end{figure}

We illustrate several training examples from digital-space training datasets, as shown in Figures~\ref{fig:RoboVQA_example}-\ref{fig:ShareRobot_example}. Specifically, RoboVQA, AgiBot, HoloAssist, and BridgeData V2 jointly provide both spatial understanding and reasoning data, enabling the model to learn physical perception and reasoning. In contrast, EgoPlan and ShareRobot primarily focus on understanding-oriented annotations, which supply perceptual grounding but contain limited reasoning requirements. Together, these datasets form a complementary corpus that supports both low-level embodied perception and higher-level reasoning capabilities.

\paragraph{RoboVQA.}
As shown in Figure~\ref{fig:RoboVQA_example}, the RoboVQA dataset provides illustrative training examples for robotic vision question answering, focusing on both spatial understanding and reasoning in digital environments. In the understanding conversation, a human queries the objects and actions in a video sequence depicting a robot arm interacting with a billiard table setup, including a dark gray felt surface, multicolored balls, and a triangular wooden rack. The AI response details the scene: the arm, white with jointed flexibility and a gripper, enters from the right, picks up a white ball, places it into the rack, and adjusts for proper positioning. The reasoning segment evaluates an agent's success in following an instruction to release the 9-number ball into the triangle, concluding failure as the white cue ball was used instead.

\paragraph{AgiBot.}
Figure ~\ref{fig:AgiBot_example} presents a training example from the AgiBot dataset, demonstrating its dual contribution to spatial understanding and reasoning capabilities. The example showcases a supermarket packing task where two robotic arms collaborate to place items into a soft woven bag. The dataset provides two types of annotations: understanding-focused descriptions that identify objects (spray bottle, mug, snack box) and their spatial relationships, and reasoning-oriented dialogues that predict next actions based on current scene observations. This complementary annotation structure enables models to develop both low-level perception skills for recognizing objects and their positions, as well as higher-level reasoning abilities for task planning and action prediction in embodied scenarios.

\paragraph{HoloAssist.}
Figure~\ref{fig:HoloAssist_example} illustrates a training example from the HoloAssist dataset, showcasing its use in training models on both spatial understanding and reasoning. The first part, "Conversations (Understanding)", demonstrates the model's ability to perceive a scene by describing objects and actions from a series of images. It correctly identifies one hand holding a black object steady while the other manipulates a small component, likely a screw. The second part, "Conversations (Reasoning)", provides the context that the current instruction is to "screw hex socket head". The model then reasons that the next logical action is to use a tool to fasten the screw, outputting the answer "insert hex socket screwdriver".

\paragraph{BridgeData V2.}
Figure \ref{fig:BridgeData_example} illustrates training examples from the BridgeData V2 dataset, a key component in digital-space corpora for embodied AI. This visualization showcases conversational annotations divided into understanding and reasoning categories. In the understanding segment, human queries prompt detailed descriptions of objects---like a silver pot, red spatula, and robotic arm---and actions in video sequences, fostering perceptual grounding in kitchen scenarios. The reasoning part involves analyzing image sequences to infer the next robotic action, such as stopping after completing a task like moving a knife beside a blue cloth.

\paragraph{EgoPlan.}
Figure~\ref{fig:EgoPlan_example} provides an illustrative example from the EgoPlan dataset, which includes a sequence of images and a conversation between a human and the GPT model. The images depict a kitchen setting, where various actions, such as moving detergent, taking a sponge, and cleaning a microwave, are being performed. The conversation between the human and the GPT model focuses on the correct sequence of actions to take based on the observed video frames. The question asked involves determining the next logical action, with the GPT model correctly identifying "clean microwave" as the next step.

\paragraph{ShareRobot.}
Figure~\ref{fig:ShareRobot_example} presents a training example from the ShareRobot dataset, showcasing a sequence of frames capturing a robotic arm's manipulation task in a kitchen environment. The images display the robot interacting with objects on a countertop, including a chips bag. The accompanying conversation demonstrates the dataset's focus on understanding-oriented annotations, where the human queries whether the robot is moving a chips bag toward the sink, to which the model responds negatively. This example illustrates ShareRobot's role in providing perceptual grounding through visual-language pairs, contributing to spatial relationship and object interaction understanding in embodied scenarios with limited reasoning requirements.

\begin{figure}[!b]
    \centering
    \includegraphics[width=0.94\textwidth]{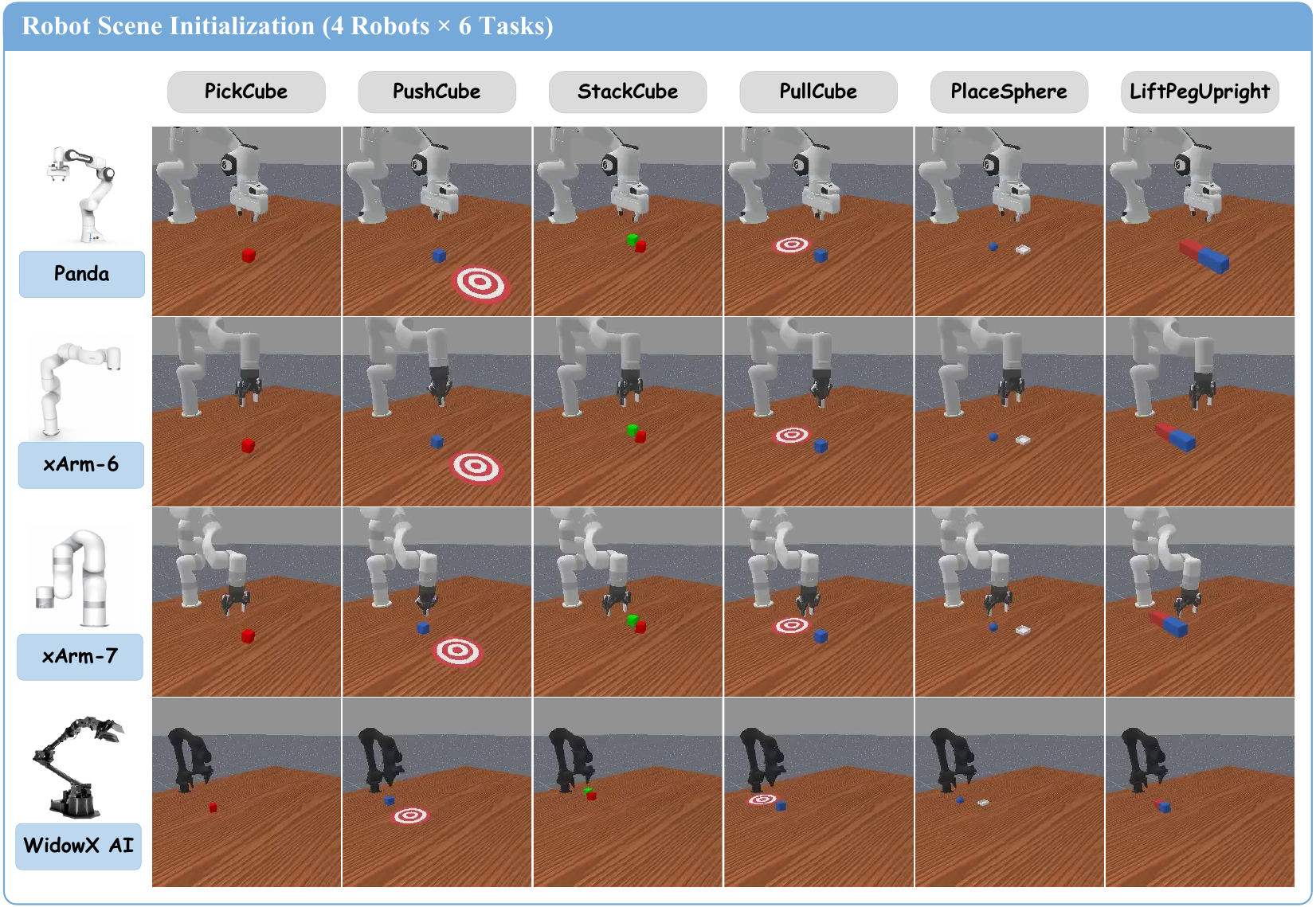}
    \vspace{3mm}
    \caption{\textbf{Visualization of robot scene initialization.}}
    \label{fig:scene_initialization}
    \vspace{-3mm}
\end{figure}
\begin{figure}[!t] 
    \centering
    \vspace{-2mm}
    \includegraphics[width=0.93\textwidth]{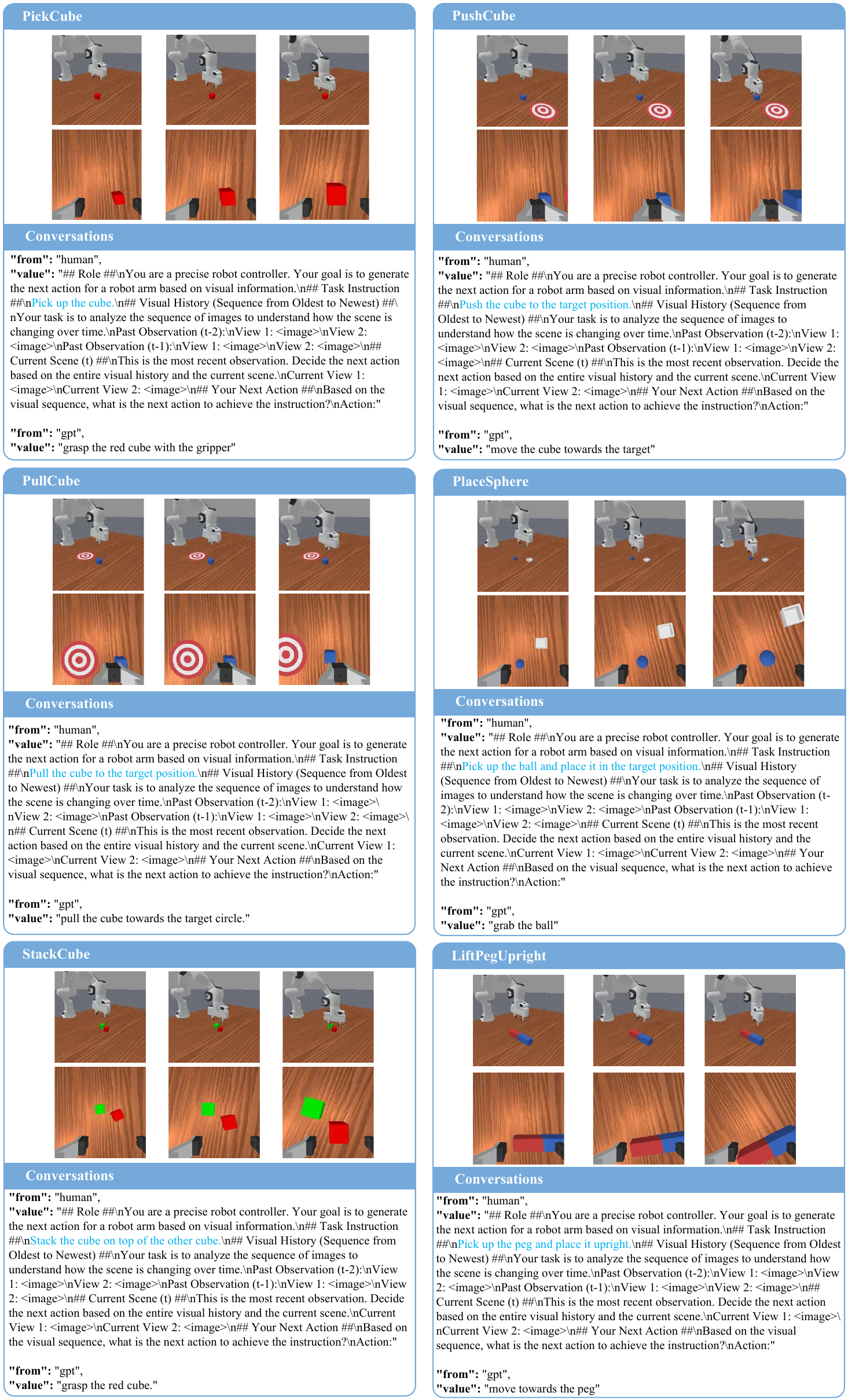}
    \vspace{5mm}
    \caption{\textbf{Visualization of ManiSkill data outputs in BLM$_1$ across six tasks.}}
    \vspace{-4mm}
    \label{fig:reasoning_sample_robot}
\end{figure}

\subsection{\textcolor{brainblue}{Examples for Physical Dataset}}

We introduce training examples from physical datasets, illustrated in Figures~\ref{fig:scene_initialization} and \ref{fig:reasoning_sample_robot}. The ManiSkill dataset, encompassing tasks such as PickCube and StackCube across Panda, xArm-6, xArm-7, and WidowX AI platforms, delivers rich data for spatial understanding and reasoning. These examples underscore the diversity of robotic morphologies and task complexities, enabling robust manipulation capabilities through integrated low-level perception and high-level reasoning. By leveraging varied robotic platforms and task scenarios, the dataset fosters comprehensive skill development, supporting advanced robotic manipulation in dynamic environments, as further detailed in the subsequent visualizations and analyses.

\paragraph{Robot Initial Scene.}
Figure \ref{fig:scene_initialization} presents the comprehensive visualization of initial scene configurations across our physical dataset, showcasing the systematic evaluation setup for four distinct robotic platforms---Panda, xArm-6, xArm-7, and WidowX AI---each tested on six manipulation tasks. The grid layout demonstrates the standardized experimental environment featuring a wooden table surface with task-specific objects including colored cubes (red, blue, green), target zones marked with circular patterns, spheres, and cylindrical pegs. This visualization highlights the diversity of both robotic morphologies and task complexities in our dataset, from simple pick-and-place operations like PickCube to more complex manipulations such as StackCube and LiftPegUpright, ensuring comprehensive coverage of common manipulation scenarios across different hardware platforms.

\paragraph{Robot Manipulation.}
Figure \ref{fig:reasoning_sample_robot} presents qualitative examples from the ManiSkill dataset showcasing BLM$_1$ performance across six robotic manipulation tasks: PickCube, PushCube, PullCube, PlaceSphere, StackCube, and LiftPegUpright. Each task demonstrates a conversational interaction where the model receives visual history sequences (observations at $t$-2, $t$-1, and current time $t$) from dual camera perspectives, along with task instructions. The model successfully generates contextually appropriate next actions based on temporal visual analysis, such as "grasp the red cube" for PickCube, "pull the cube towards the target circle" for PullCube, and "move towards the peg" for LiftPegUpright, demonstrating robust visual understanding and action planning capabilities in diverse scenarios.

\section{\textcolor{brainblue}{Implementation Details}}
\label{sec:appendix_implementation}

\subsection{\textcolor{brainblue}{Hyper-parameters of Stage I}}  
In Stage I, BLM$_1$ is fine-tuned using pretrained MLLMs with multimodal data, as shown in Table~\ref{tab:training_setup}. Key hyperparameters include a maximum token length of 4,096, pixel length range of 12,544 to 451,584, and a peak learning rate of 1e-6 with AdamW optimizer ($\beta_1$=0.9, $\beta_2$=0.999, $\epsilon$=1e-8). The cosine decay schedule and batch size of 128 ensure stable training over 18,046 steps, preserving general reasoning.

\input{Tables/appendix_hyperparameter}

\subsection{\textcolor{brainblue}{Hyper-parameters of Stage II}}  
Stage II employs cross-embodiment learning with flow-matching loss, as detailed in Table~\ref{tab:Stage_II_training_setup}. Using a Perceiver architecture, a peak learning rate of 1e-4, and a batch size of 128, the model trains for 200,000 steps with cosine decay. The DiT model, with hidden dimensions of 1,536 and 16 layers, uses a noise scheduling ($\alpha$=1.5, $\beta$=1.0) to handle diverse robotic embodiments effectively.

\input{Tables/appendix_hyperparameter_2}

\subsection{\textcolor{brainblue}{Hyper-parameters of Robotic Baselines}}
In Stage II, the robotic baselines are trained with different models, with specific configurations detailed in Table \ref{tab:baseline_setup}. The $\pi_0$ model utilizes a batch size of 32, with a peak learning rate of 2e-5, decaying to 1.5e-6 over 100,000 steps. The HPT model is trained for 1,000 epochs with a batch size of 768, while UniAct uses a batch size of 64, a peak learning rate of 1e-4, and 1.5 million training steps. The GR00T-N1 and GR00T-N1.5 models, both trained for 300,000 steps, share similar configurations with a batch size of 32 and 128, respectively, and a peak learning rate of 1e-4 with cosine decay.

\input{Tables/hyper-parameters_stage2_baselines}

\section{\textcolor{brainblue}{Prompt Details}}
\begin{figure}[!h]
    \centering
    \includegraphics[width=1\textwidth]{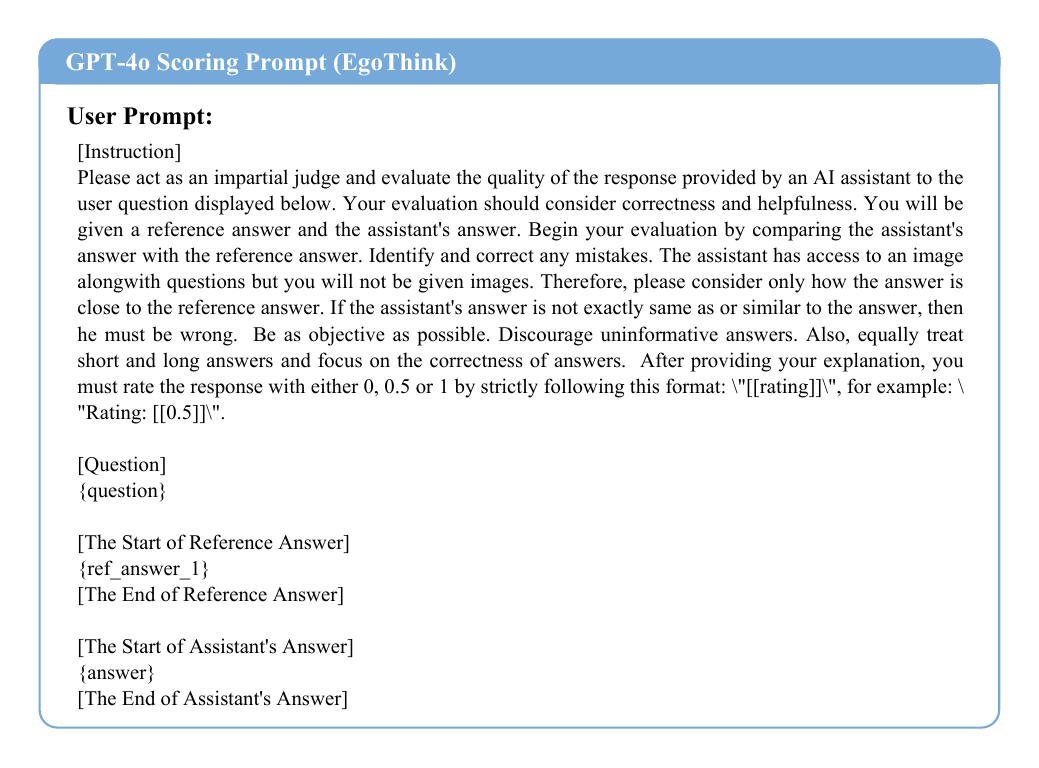}
    \caption{\textbf{GPT-4o prompt for EgoThink.}}
    \label{fig:Prompt_GPT4o}
\end{figure}

This section systematically presents all prompts used for model inference and scoring in the evaluation phase. Figure~\ref{fig:Prompt_MCQ} shows the inference templates for the RoboVQA, AgiBot, HoloAssist, and RoboFail benchmarks. Figures~\ref{fig:Prompt_QA_EgoThink} and~\ref{fig:Prompt_QA_ShareRobot} display the free-form inference prompts for EgoThink and ShareRobot, respectively. RoboVQA, AgiBot, HoloAssist, and RoboFail employ a multiple-choice format, with performance measured by accuracy. Conversely, EgoThink and ShareRobot use a free-form paradigm, assessed via an LLM-as-a-Judge protocol where a large language model scores responses. Figures~\ref{fig:Prompt_GPT4o} and~\ref{fig:Prompt_GPT4V} illustrate the scoring prompts for these benchmarks.

\subsection{\textcolor{brainblue}{Inference Prompts}}
\begin{figure}[!b]
    \vspace{-1em}   
    \centering
    \includegraphics[width=1\textwidth]{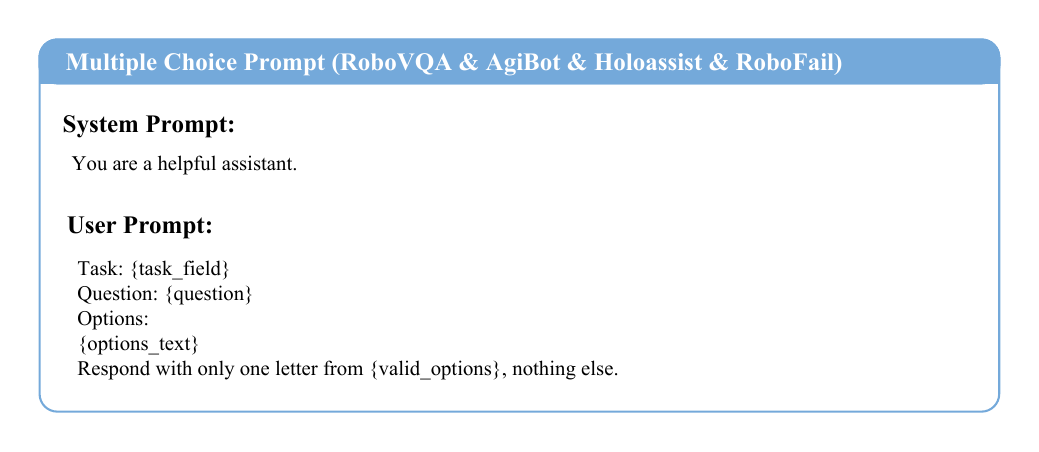}
    \caption{\textbf{Multiple-choice question prompt for RoboVQA, AgiBot, HoloAssist, and RoboFail.}}
    \label{fig:Prompt_MCQ}
\end{figure}

\paragraph{Multiple-Choice Prompt.}
As shown in Figure~\ref{fig:Prompt_MCQ}, the multiple-choice question (MCQ) prompt for RoboVQA, AgiBot, HoloAssist, and RoboFail evaluates robotic task performance. The prompt includes placeholders: \texttt{\{task\_field\}} specifies the task context, such as success or affordance evaluation; \texttt{\{question\}} formulates the specific query about the agent's action or next subtask; and \texttt{\{options\_text\}} lists the answer choices (e.g., A, B, C, D) with corresponding responses. 
\begin{figure}[!b]
    \vspace{-1em}   
    \centering
    \includegraphics[width=1\textwidth]{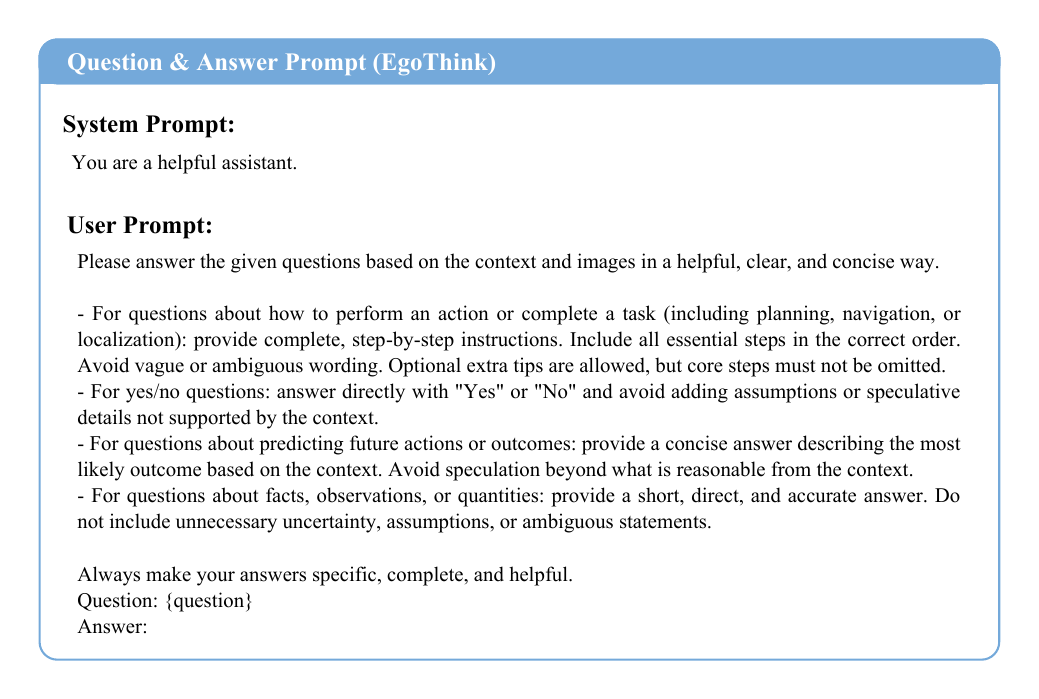}
    \caption{\textbf{Free-formed QA prompt for EgoThink.}}
    \label{fig:Prompt_QA_EgoThink}
\end{figure}

\paragraph{EgoThink QA Prompt.}
Figure~\ref{fig:Prompt_QA_EgoThink} presents the question-answering prompt template for EgoThink. The system prompt establishes a helpful assistant role, while the user prompt provides structured response guidelines. The \texttt{question} placeholder is dynamically replaced with the actual query during inference. The prompt design enforces specific answer formats: step-by-step instructions for procedural tasks, binary responses for yes/no questions, predictive descriptions for future outcomes, and concise factual statements for observational queries.

\begin{figure}[!h]
    \centering
    \includegraphics[width=1\textwidth]{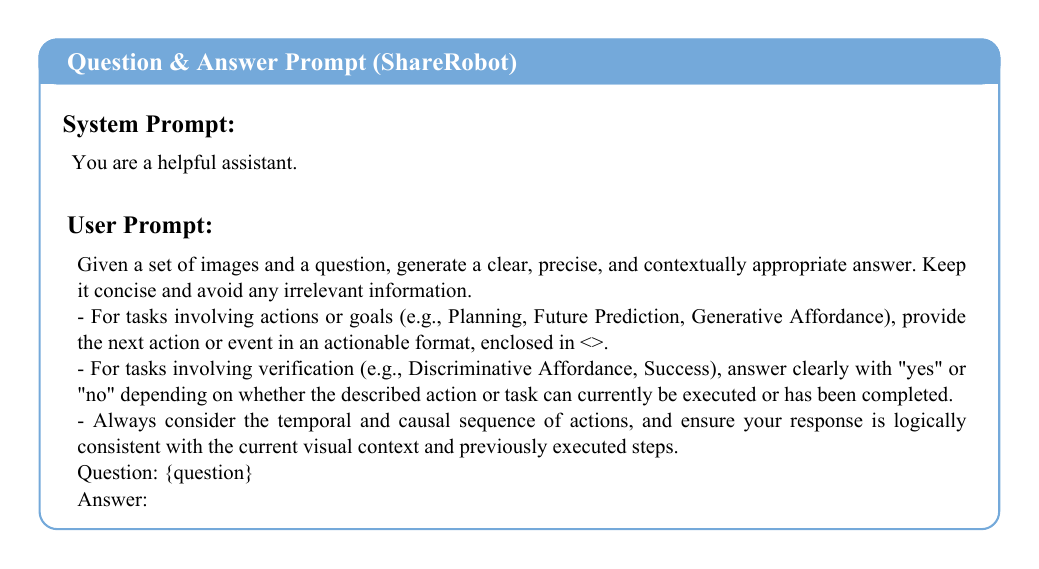}
    \caption{\textbf{Free-formed QA prompt for ShareRobot.}}
    \label{fig:Prompt_QA_ShareRobot}
\end{figure}

\paragraph{ShareRobot QA Prompt.}
As illustrated in Figure~\ref{fig:Prompt_QA_ShareRobot}, the QA prompt for ShareRobot consists of a simple system prompt and a detailed user prompt. The user prompt incorporates the placeholder \texttt{\{question\}}, which is substituted with the specific query posed to the model. This design emphasizes concise, contextually relevant responses, with specialized formats for action-oriented tasks (enclosed in \texttt{<>}) and verification queries ("yes" or "no"), while maintaining temporal and causal logic aligned with visual inputs.

\subsection{\textcolor{brainblue}{Scoring Prompts}} \label{sec:scoring_prompts}
\paragraph{GPT-4o Scoring Prompt.}
The GPT-4o scoring prompt for EgoThink, illustrated in Figure~\ref{fig:Prompt_GPT4o}, directs the AI to serve as an impartial evaluator, assessing assistant responses for correctness and helpfulness by comparing them to reference answers. It incorporates placeholders: \texttt{\{question\}} for the user query, \texttt{\{ref\_answer\_1\}} for the reference answer, and \texttt{\{answer\}} for the assistant's response. The scoring mechanism mandates an explanation followed by a rating of 0, 0.5, or 1 in the precise format "Rating: [[rating]]", such as "Rating: [[0.5]]".

\begin{figure}[!h]
    \centering
    \includegraphics[width=1\textwidth]{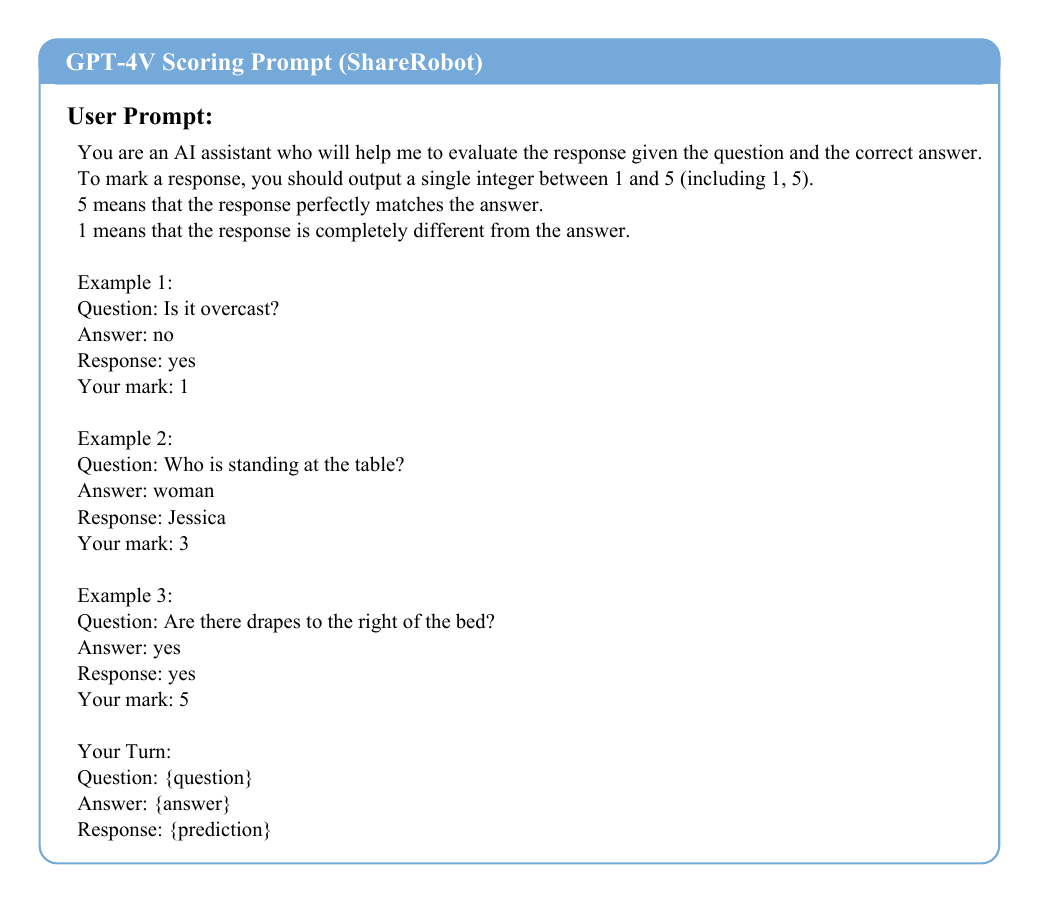}
    \caption{\textbf{GPT-4V prompt for ShareRobot.}}
    \label{fig:Prompt_GPT4V}
\end{figure}

\paragraph{GPT-4V Scoring Prompt.}
As shown in Figure~\ref{fig:Prompt_GPT4V}, the prompt evaluates model responses using placeholders: \texttt{\{question\}} for the query, \texttt{\{answer\}} for the ground truth, and \texttt{\{prediction\}} for the model's output. The scoring mechanism instructs outputting a single integer from 1 to 5, where 5 indicates a perfect match with the answer and 1 signifies complete dissimilarity. Examples illustrate grading nuances, ensuring consistent evaluation norms for visual question answering tasks.
\end{document}

%% file: Tables/A_summary_of_datasets_used_for_BLM-1_stage_1.tex
\begin{table*}[!b]
    \caption{\textbf{Summary of the dataset recipe for BLM$_1$-Stage I training.}} 
    \label{table:digital_statistics_train}
    \renewcommand{\arraystretch}{1.2}
    \setlength{\tabcolsep}{3mm}
    \centering
    \resizebox{\textwidth}{!}{
    \begin{tabular}{c|cccc|cc|c}
        \Xhline{0.7pt}
        \specialrule{0em}{0pt}{0pt}
        \textbf{Type} & \textbf{RoboVQA} & \textbf{AgiBot} & \textbf{HoloAssist} & \textbf{BridgeData V2} & \textbf{EgoPlan} & \textbf{ShareRobot} & \textbf{Total}\\
        \hline
        Understanding & 218.5K & 19.4K & 136.3K & 129.2K & \multirow{2}{*}{52.1K} & \multirow{2}{*}{1.0M} & \multirow{2}{*}{2.8M} \\
        Reasoning & 920.0K & 19.4K & 136.3K & 129.1K & & & \\
        \Xhline{0.7pt}
    \end{tabular}
    }
\end{table*}

%% file: Tables/A_summary_of_benchmarks_used_for_BLM-1.tex
\begin{table*}[!b]
    \caption{\textbf{Summary of benchmarks used for BLM$_1$.}}
    \label{table:digital_statistics_test}
    \renewcommand{\arraystretch}{1.2}
    \setlength{\tabcolsep}{4.75mm}
    \centering
    \scalebox{0.75}{ 
    \begin{tabular}{c|cccccc|c}
        \Xhline{0.7pt}
        \specialrule{0em}{0pt}{0pt}
        \textbf{Dataset} & \textbf{RoboVQA} & \textbf{AgiBot} & \textbf{HoloAssist} & \textbf{RoboFail} & \textbf{EgoThink} & \textbf{ShareRobot} & \textbf{Total}\\
        \hline
        \multicolumn{1}{c|}{Test} & 110 & 100 & 100 & 100 & 700 & 2,050 & 3,160 \\
        \Xhline{0.7pt}
    \end{tabular}
    }
\end{table*}

%% file: Tables/A_summary_of_datasets_used_for_BLM-1_stage_2.tex
\begin{table*}[b]
    \caption{\textbf{Summary of the dataset recipe for BLM$_1$-Stage II training.}} 
    \label{table:physical_statistics_train}
    \renewcommand{\arraystretch}{1.2}
    \setlength{\tabcolsep}{3.75mm}
    \centering
    \scalebox{0.75}{ 
    \begin{tabular}{c|cccccc|c}
        \Xhline{0.7pt}
        \specialrule{0em}{0pt}{0pt}
        \textbf{Type}  & \textbf{PickCube} & \textbf{PullCube} & \textbf{StackCube} & \textbf{PushCube} & \textbf{PlaceSphere} & \textbf{LiftPegUpright} & \textbf{Total}\\
        \hline
        \multicolumn{1}{c|}{Physical} & 31.4K  & 39.9K & 52.6K & 98.6K & 49.1K & 76.2K & 347.8K \\ 
        \Xhline{0.7pt}
    \end{tabular}
    \label{tab:stage2_dataset_summary}
    }
\end{table*}

%% file: Tables/comparison_existing_MLLMs_and_GMLMs_on_spatial_reasoning_benchmarks.tex
\begin{table*}[b]
    \caption{\textbf{Comparison with existing closed/open-source MLLMs, embodied large language models and general multimodal large models on digital-space benchmarks.}} 
    \label{table:digital-space benchmarks}
    \renewcommand{\arraystretch}{1.2}
    \setlength{\tabcolsep}{2.84mm}
    \centering
    \scalebox{0.75}{ 
    \begin{tabular}{l|cccccc|c}
        \Xhline{0.7pt}
        \multicolumn{1}{l|}{\multirow{2}{*}{\textbf{Models}}} & \multicolumn{6}{c|}{\textbf{Benchmarks}} & \multicolumn{1}{c}{\multirow{2}{*}{\textbf{Avg. $\uparrow$}}}\\ \cline{2-7}
        \specialrule{0em}{0pt}{0pt}
        & \textbf{RoboVQA}  & \textbf{AgiBot} & \textbf{HoloAssist} & \textbf{RoboFail} & \textbf{EgoThink} & \textbf{ShareRobot}\\
        \hline
        \multicolumn{8}{l}{\cellcolor{gray!20} \raisebox{0.2em}[0pt][0pt]{$\downtriangle$} \raisebox{-0.2em}[0pt][0pt]{\textbf{Closed-source MLLMs}}} \\ [2pt] 
        \multicolumn{1}{l|}{GPT-4o \cite{MLLM:GPT4o}} & 75.45 & 38.00 & \underline{54.00} & 58.00 & \textbf{72.42} & \underline{61.30} & \underline{59.86}\\ 
        \multicolumn{1}{l|}{Claude-3.5 Sonnet \cite{MLLM:Claude3.5Sonnet}} & 64.55 & 32.00 & 46.00 & 58.00 & 48.54 & 53.20 & 50.38\\ 
        \hhline{-|-------}
        \multicolumn{8}{l}{\cellcolor{gray!20}\raisebox{0.2em}[0pt][0pt]{$\downtriangle$} \raisebox{-0.2em}[0pt][0pt]{\textbf{Open-source MLLMs}}} \\ [2pt] 
        \multicolumn{1}{l|}{LLaVA-OneVision-7B \cite{MLLM:LlavaOneVision}} & 75.45 & 29.00 & 31.00 & 58.00 & 59.46 & 40.41 & 48.89 \\ 
        \multicolumn{1}{l|}{InternVL2.5-8B \cite{MLLM:InternVL}} & 69.09 & 29.00 & 37.00 & 60.00 & \underline{61.54} & 45.06& 50.28 \\ 
        \multicolumn{1}{l|}{Qwen2.5-VL-7B-Instruct \cite{MLLM:Qwen2.5-VL}}  & 86.36 & \underline{42.00} & 49.00 & \underline{64.00} & 57.00 & 46.67 & 57.51 \\ 
        \multicolumn{1}{l|}{Cosmos-7B \cite{ELLM:Cosmos-Reason1}}  &  \underline{90.91} & 39.00 & 45.00 & 62.00 & 56.67 & 57.70 & 58.55\\
        \hhline{-|-------}
        \multicolumn{8}{l}{\cellcolor{gray!20} \raisebox{0.2em}[0pt][0pt]{$\downtriangle$} \raisebox{-0.2em}[0pt][0pt]{\textbf{Embodied Large Language Models}}} \\ [2pt]
        \multicolumn{1}{l|}{Magma-8B \cite{ELLM:Magma}} & 47.27  & 28.00 & 35.00 & 49.00 &48.83 &35.17 &40.55 \\  
        \multicolumn{1}{l|}{VeBrain-7B \cite{ELLM:VeBrain}} & 85.45 & 41.00 & 44.00 & 55.00 & 59.33 & 51.28 & 56.01\\ 
        \hhline{-|-------}
        \multicolumn{8}{l}{\cellcolor{gray!20} \raisebox{0.2em}[0pt][0pt]{$\downtriangle$} \raisebox{-0.2em}[0pt][0pt]{\textbf{General Multimodal Large Models}}} \\ [2pt] 
        \multicolumn{1}{l|}{ChatVLA-2B \cite{GMLM:ChatVLA}} & 60.91 & 38.00 & 36.00 & 52.00 & 47.71& 35.80&45.07 \\ 
        \multicolumn{1}{l|}{RoboBrain2-7B \cite{ELLM:RoboBrain}} & 72.73 & 31.00 & 48.00 & 54.00 & 61.21& 51.88&53.14 \\ 
        \hhline{-|-------}
        \multicolumn{1}{l|}{\textbf{BLM$_1$}} & \textbf{95.45} & \textbf{47.00} & \textbf{56.00} & \textbf{66.00} & 59.42 & \textbf{65.40} & \textbf{64.88} \\ 
        \Xhline{0.7pt}
    \end{tabular}
    }
\end{table*}

%% file: Tables/Comparison_with_existing_MLLMs_and_GMLMs_on_EgoThink.tex
\begin{table*}[t]
    \caption{\textbf{Comparison with existing MLLMs, ELLMs and GMLMs on EgoThink.}} 
    \label{EgoThink}
    \renewcommand{\arraystretch}{1.2}
    \setlength{\tabcolsep}{0.7mm}
    \centering
    \scalebox{0.577}{ 
    \begin{tabular}{l|cccccccccccc|c}
        \Xhline{0.7pt}
        \multicolumn{1}{l|}{\multirow{2}{*}{\textbf{Models}}} & \multicolumn{1}{c|}{\multirow{2}{*}{\textbf{Activity}}} & \multicolumn{1}{c|}{\multirow{2}{*}{\textbf{Forecast}}} & \multicolumn{2}{c|}{\textbf{Localization}}& \multicolumn{3}{c|}{\textbf{Object}}& \multicolumn{3}{c|}{\textbf{Planning}} & \multicolumn{2}{c|}{\textbf{Reasoning}} &  \multicolumn{1}{c}{\multirow{2}{*}{\textbf{Avg. $\uparrow$}}}\\ 
        \cline{4-5} \cline{6-8} \cline{9-11} \cline{12-13}
        \specialrule{0em}{0pt}{0pt}
          & \multicolumn{1}{c|}{} & \multicolumn{1}{c|}{} & \textbf{Location}  & \multicolumn{1}{c|}{\textbf{Spatial}} & \textbf{Affordance} & \textbf{Attribute} & \multicolumn{1}{c|}{\textbf{Existence}} & \textbf{Assistance} & \textbf{Navigation} & \multicolumn{1}{c|}{\textbf{Comparing}} & \textbf{Counting} & \textbf{Situated} \\
        \hline
        \multicolumn{14}{l}{\cellcolor{gray!20} \raisebox{0.2em}[0pt][0pt]{$\downtriangle$} \raisebox{-0.2em}[0pt][0pt]{\textbf{Closed-source MLLMs}}} \\ [2pt] 
         \multicolumn{1}{l|}{GPT-4o ~\cite{MLLM:GPT4o}} & 70.50 & 69.50 & 91.00 & 74.00 & 74.00 & 88.00 & 71.00 & 73.00 & 32.00 & 75.00 & 68.00 & 83.00 & 72.42\\ 
         \multicolumn{1}{l|}{Claude-3.5 Sonnet ~\cite{MLLM:Claude3.5Sonnet}} & 46.00 & 44.50 & 80.00 & 38.00 & 46.00 & 63.00 & 48.00 & 60.00 & 11.00 & 49.00 & 38.00 & 59.00 & 48.54 \\
        \hhline{-|-------------}
         \multicolumn{14}{l}{\cellcolor{gray!20}\raisebox{0.2em}[0pt][0pt]{$\downtriangle$} \raisebox{-0.2em}[0pt][0pt]{\textbf{Open-source MLLMs}}} \\ [2pt] 
         \multicolumn{1}{l|}{LLaVA-One-Vision-7B ~\cite{MLLM:LlavaOneVision}} & 51.50 & 58.00 & 90.00 & 64.00 & 58.00 & 74.00 & 69.00 & 47.00 & 15.00 & 69.00 & 46.00 & 72.00 & 59.46\\ 
         \multicolumn{1}{l|}{InternVL2.5-8B ~\cite{MLLM:InternVL}} & 60.50 & 56.00 & 87.00 & 68.00 & 66.00 & 72.00 & 71.00 & 54.00 & 27.00 & 64.00 & 58.00 & 55.00 & 61.54\\ 
         \multicolumn{1}{l|}{Qwen2.5-VL-7B-Instruct ~\cite{MLLM:Qwen2.5-VL}} & 58.00 & 50.00 & 74.00 & 79.00 & 61.00 & 65.00 & 62.00 & 54.00 & 20.00 & 60.00 & 48.00 & 53.00 & 57.00 \\ 
         \multicolumn{1}{l|}{Cosmos-7B ~\cite{ELLM:Cosmos-Reason1}}  & 54.50 & 47.50 & 84.00 & 68.00 & 55.00 & 65.00 & 62.00 & 52.00 & 18.00 & 63.00 & 48.00 & 63.00 & 56.67  \\
         \hhline{-|-------------}
        \multicolumn{14}{l}{\cellcolor{gray!20} \raisebox{0.2em}[0pt][0pt]{$\downtriangle$} \textbf{Embodied Large Language Models}} \\ [2pt]
         \multicolumn{1}{l|}{Magma-8B ~\cite{ELLM:Magma}} & 32.00 & 52.00 & 78.00 & 70.00 & 58.00 & 78.00 & 62.00 & 9.00 & 9.00 & 58.00 & 34.00 & 46.00 & 48.83 \\ 
         \multicolumn{1}{l|}{VeBrain-7B ~\cite{ELLM:VeBrain}} & 63.00 & 56.00 & 90.00 & 67.00 & 64.00 & 68.00 & 73.00 & 39.00 & 13.00 & 60.00 & 58.00 & 61.00 & 59.33 \\ 
        \hhline{-|-------------}
        \multicolumn{14}{l}{\cellcolor{gray!20} \raisebox{0.2em}[0pt][0pt]{$\downtriangle$} \raisebox{-0.2em}[0pt][0pt]{\textbf{General Multimodal Large Models}}} \\ [2pt] 
         \multicolumn{1}{l|}{ChatVLA-2B ~\cite{GMLM:ChatVLA}} & 46.50 & 47.00 & 75.00 & 64.00 & 44.00 & 69.00 & 58.00 & 20.00 & 8.00 & 40.00 & 36.00 & 65.00 & 47.71 \\ 
         \multicolumn{1}{l|}{RoboBrain2-7B ~\cite{ELLM:RoboBrain}} & 53.00 & 46.50 & 93.00 & 70.00 & 75.00 & 79.00 & 77.00 & 49.00 & 18.00 & 62.00 & 44.00 & 68.00 & 61.21 \\ 

         \hhline{-|-------------}
         \multicolumn{1}{l|}{\textbf{BLM$_{1}$}} & 55.00 & 51.00 & 81.00 & 71.00 & 61.00 & 76.00 & 63.00 & 55.00 & 25.00 & 60.00 & 56.00 & 59.00 & 59.42 \\ 
        \Xhline{0.7pt}
    \end{tabular}
    }
\end{table*}

%% file: Tables/Comparison_existing_MLLMs_and_GMLMs_on_ShareRobot.tex
\begin{table*}[t]
    \caption{\textbf{Comparison with existing MLLMs, ELLMs and GMLMs on ShareRobot.}}
    \label{ShareRobot}
    \renewcommand{\arraystretch}{1.2}
    \setlength{\tabcolsep}{0.8mm}
    \centering
    \scalebox{0.588}{
    \begin{tabular}{l|cccccccccc|c}
        \Xhline{0.7pt}
        \multicolumn{1}{l|}{\multirow{4}{*}{\textbf{Models}}} & \multicolumn{10}{c|}{\textbf{Task Category}} & \multicolumn{1}{c}{\multirow{4}{*}{\textbf{Avg. $\uparrow$}}}\\
        \cline{2-11}
        & \textbf{\makecell{Discriminative\\ Affordance \\ (Negative)}} & \textbf{\makecell{Discriminative \\ Affordance \\ (Positive)}} & \textbf{\makecell{Future \\ Prediction}} & \textbf{\makecell{Generative \\ Affordance}} & \textbf{\makecell{ Past \\ Description}} & \textbf{\makecell{Planning \\ Remaining \\ Steps}} & \textbf{\makecell{ Planning}} &  \textbf{\makecell{ Planning \\ with \\ Context}} & \textbf{\makecell{ Success \\ (Negative)}}&  \textbf{\makecell{ Success \\ (Positive)}} \\
        \hline
        \multicolumn{12}{l}{\cellcolor{gray!20} \raisebox{0.2em}[0pt][0pt]{$\downtriangle$} \textbf{ Closed-source MLLMs}} \\ [2pt]
        \multicolumn{1}{l|}{GPT-4o ~\cite{MLLM:GPT4o}} & 95.61 & 54.15 & 63.17 & 29.02 & 33.66 & 75.49 & 62.80 & 86.46 & 36.59 & 76.10 & 61.30 \\
        \multicolumn{1}{l|}{Claude-3.5 Sonnet \cite{MLLM:Claude3.5Sonnet}} & 65.73 & 65.24 & 50.49 & 23.05 & 15.12 & 68.78 & 60.98 & 75.73 & 36.46 & 70.37 & 53.20 \\
        \hhline{-|-----------}
        \multicolumn{12}{l}{\cellcolor{gray!20} \raisebox{0.2em}[0pt][0pt]{$\downtriangle$} \textbf{ Open-source MLLMs}} \\ [2pt]
        \multicolumn{1}{l|}{LLaVA-One-Vision-7B \cite{MLLM:LlavaOneVision}} & 69.27 & 72.20 & 29.88 & 13.90 & 11.22 & 26.71 & 40.00 & 44.39 & 38.05 & 58.54 & 40.41\\
        \multicolumn{1}{l|}{InternVL2.5-8B \cite{MLLM:InternVL}} & 80.73 & 62.56 & 34.76 & 9.76 & 13.90 & 38.90 & 47.68 & 56.95 & 41.83 & 63.54 & 45.06\\
        \multicolumn{1}{l|}{Qwen2.5-VL-7B-Instruct \cite{MLLM:Qwen2.5-VL}}  & 99.02 & 32.20 & 50.12 & 13.90 & 21.22 & 41.46 & 46.22 & 56.22 & 93.17 & 13.17 & 46.67\\
        \multicolumn{1}{l|}{Cosmos-7B \cite{ELLM:Cosmos-Reason1}}  & 93.17 & 42.93 & 70.37 & 25.12 & 27.93 & 64.39 & 61.34 & 82.32 & 48.29 & 60.98 & 57.68 \\
        \hhline{-|-----------}
        \multicolumn{12}{l}{\cellcolor{gray!20} \raisebox{0.2em}[0pt][0pt]{$\downtriangle$} \textbf{Embodied Large Language Models}} \\ [2pt]
        \multicolumn{1}{l|}{Magma-8B \cite{ELLM:Magma}} & 100.00 & 10.24 & 39.39 & 3.66 & 5.37 & 33.05 & 26.59 & 32.44 & 96.10 & 4.88 & 35.17 \\
        \multicolumn{1}{l|}{VeBrain-7B \cite{ELLM:VeBrain}} & 84.88 & 80.49 & 52.20 & 17.68 & 19.27 & 47.20 & 40.00 & 57.44 & 53.66 & 60.00 & 51.28 \\
        \hhline{-|-----------}
        \multicolumn{12}{l}{\cellcolor{gray!20} \raisebox{0.2em}[0pt][0pt]{$\downtriangle$} \textbf{ General Multimodal Large Models}} \\ [2pt]
        \multicolumn{1}{l|}{ChatVLA-2B \cite{GMLM:ChatVLA}} & 87.32 & 57.56 & 27.20 & 11.46 & 9.63 & 0.24 & 45.37 & 25.12 & 43.90 & 50.24 & 35.80 \\ 
        \multicolumn{1}{l|}{RoboBrain2-7B \cite{ELLM:RoboBrain}} & 65.12 & 77.07 & 58.78 & 22.07 & 22.44 & 41.59 & 47.32 & 69.27 & 40.98 & 74.15 & 51.88 \\
        \hhline{-|-----------}
        \multicolumn{1}{l|}{\textbf{BLM$_1$}} & 94.63 & 84.88 & 73.29 & 29.15 & 30.24 & 72.80 & 60.49 & 85.73 & 51.71 & 71.22 & 65.41\\
        \Xhline{0.7pt}
    \end{tabular}
    }
\end{table*}

%% file: Tables/comparison_existing_VLAs_on_robot_benchmarks2.tex
\begin{table}[b]
    \caption{\textbf{Comparison with existing VLAs on robot benchmarks.} \textsuperscript{$\dagger$} denotes the training of independent models on four robots, with each model evaluated across six tasks. \textsuperscript{$\star$} denotes training independent models for each of the six tasks associated with four robots (24 models in total),  with evaluation on the corresponding tasks for each robot.} 
    \vspace{3.5mm}
    \label{spatial_reasoning2}
    \renewcommand{\arraystretch}{1.2}
    \setlength{\tabcolsep}{7.4mm}
    \centering
    \scalebox{0.75}{ 
    \begin{tabular}{l|cccc|c}
        \Xhline{0.7pt}
        \textbf{Models} & \textbf{Panda} &  \textbf{xArm-6} & \textbf{xArm-7} & \textbf{WidowX AI} & \textbf{Avg.\,$\uparrow$}\\ 
        \hline
        \multicolumn{6}{l}{\cellcolor{gray!20} \raisebox{0.2em}[0pt][0pt]{\downtriangle} \raisebox{-0.2em}[0pt][0pt]{\textbf{Pre-trained Models}}} \\ [2pt] 
         \multicolumn{1}{l|}{{\Large $\pi$}\textsubscript{0}~\cite{VLA:Pi0}} & \underline{84.00\%} & 67.33\% & 55.33\% & 63.00\% & 67.42\%\\ 
         \multicolumn{1}{l|}{{\Large $\pi$}\textsubscript{0}\textsuperscript{$\dagger$}~\cite{VLA:Pi0}} & 82.33\% & 66.33\% & 60.00\% & 55.67\% & 66.08\%\\  
         \multicolumn{1}{l|}{HPT~\cite{VLA:HPT}} & 55.00\% & 48.67\% & 51.33\% & 45.00\% & 50.00\% \\ 
         \multicolumn{1}{l|}{UniAct~\cite{VLA:UniACT}} & 51.67\% & 52.67\% & 51.33\% & 19.00\% & 43.67\% \\ 
         \multicolumn{1}{l|}{GR00T-N1~\cite{VLA:GR00T_N1}} & 67.67\% & 69.33\% & 64.00\% & 59.33\% & 65.08\% \\ 
         \multicolumn{1}{l|}{GR00T-N1.5~\cite{GMLM:GR00T_N15}} & 80.33\% & \textbf{84.67\%} & 71.33\% & \textbf{65.33\%} & \underline{75.42\%} \\ 
        \hhline{-|-----}
         \multicolumn{6}{l}{\cellcolor{gray!20}\raisebox{0.2em}[0pt][0pt]{\downtriangle} \raisebox{-0.2em}[0pt][0pt]{\textbf{From-scratch Models}}} \\ [2pt] 
         \multicolumn{1}{l|}{Diffusion Policy~\cite{VLA:DP}} & 51.67\% & 64.00\% & 66.33\% & 40.00\% & 55.50\%  \\
         \multicolumn{1}{l|}{Diffusion Policy\textsuperscript{$\dagger$}~\cite{VLA:DP}} & 66.00\% & 69.00\% & 69.33\% & 52.00\% & 64.08\%  \\ 
         \multicolumn{1}{l|}{Diffusion Policy\textsuperscript{$\star$}~\cite{VLA:DP}} & 74.00\% & 79.33\% & \textbf{78.33\%} & 55.00\% & 71.67\%  \\ 
         \multicolumn{1}{l|}{GR00T-N1.5~\cite{GMLM:GR00T_N15}} & 82.67\% & 74.33\% & 70.67\% & 63.33\% & 72.75\%\\ 
         \hhline{-|-----}
         \multicolumn{1}{l|}{\textbf{BLM$_1$}} & \textbf{84.67\%} & \underline{80.33\%} & \underline{74.67\%} & \underline{63.67\%} & \textbf{75.83\%}  \\ 
        \Xhline{0.7pt}
    \end{tabular}
    \label{tab:physical_space}
    }
\end{table}

%% file: Tables/comparison_existing_VLAs_on_robot_benchmarks.tex
\begin{table*}[t]
    \caption{\textbf{Comparison with existing VLAs on robot benchmarks.} \textsuperscript{$\dagger$} denotes the training of independent models on four robots, with each model evaluated across six tasks. \textsuperscript{$\star$} denotes training independent models for each of the six tasks associated with four robots (24 models in total),  with evaluation on the corresponding tasks for each robot.} 
    \label{spatial_reasoning}
    \renewcommand{\arraystretch}{1.2}
    \setlength{\tabcolsep}{1.2mm}
    \centering
    \scalebox{0.685}{ 
    \begin{tabular}{l|c|cccccc|c|c}
        \Xhline{0.7pt}
        \textbf{Models} & \textbf{Robot} &\textbf{PickCube} &  \textbf{PushCube} & \textbf{StackCube} & \textbf{PullCube} & \textbf{PlaceSphere} & \textbf{LiftPegUpright} & \textbf{Robot Avg.} & \textbf{Avg.\,$\uparrow$}\\  
        \hline
        \multicolumn{10}{l}{\cellcolor{gray!20} \raisebox{0.2em}[0pt][0pt]{\downtriangle} \raisebox{-0.2em}[0pt][0pt]{\textbf{Pre-trained Models}}} \\ [2pt] 
         \multicolumn{1}{l|}{\multirow{4}{*}{{\Large $\pi$}\textsubscript{0}~\cite{VLA:Pi0}}} & Panda & 98.00\% & 96.00\% & 74.00\% & 82.00\% & 94.00\% & 60.00\% & 84.00\% & \multicolumn{1}{c}{\multirow{4}{*}{67.42\%}} \\ 
          & xArm-6 & 78.00\% & 92.00\% & 46.00\% & 90.00\% & 38.00\% & 60.00\% & 67.33\% &\\ 
          & xArm-7 & 58.00\% & 82.00\% & 36.00\% & 90.00\% & 28.00\% & 38.00\% & 55.33\% & \\ 
          & WidowX AI & 100.00\% & 50.00\% & 26.00\% & 54.00\% & 78.00\% & 70.00\% &  63.00\% & \\ 
          \hhline{-|---------}
         \multicolumn{1}{l|}{\multirow{4}{*}{{\Large $\pi$}\textsubscript{0}\textsuperscript{$\dagger$}~\cite{VLA:Pi0}}} & Panda & 100.00\% & 96.00\% & 80.00\% & 82.00\% & 82.00\% & 54.00\% & 82.33\% & \multicolumn{1}{c}{\multirow{4}{*}{66.08\%}} \\ 
          & xArm-6 & 56.00\% & 94.00\% & 56.00\% & 98.00\% & 58.00\% & 36.00\% & 66.33\% &\\ 
          & xArm-7 & 50.00\% & 86.00\% & 24.00\% & 98.00\% & 46.00\% & 56.00\% & 60.00\% & \\ 
          & WidowX AI & 100.00\% & 54.00\% & 14.00\% & 44.00\% & 52.00\% & 70.00\% & 55.67\% & \\ 
         \hhline{-|---------}
         \multicolumn{1}{l|}{\multirow{4}{*}{{HPT}~\cite{VLA:HPT}}} & Panda & 90.00\% & 94.00\% & 0.00\% & 88.00\% & 18.00\% & 40.00\% & 55.00\% & \multicolumn{1}{c}{\multirow{4}{*}{50.00\%}} \\ 
          & xArm-6 & 68.00\% & 64.00\% & 6.00\% & 60.00\% & 32.00\% & 62.00\% & 48.67\% &\\ 		
          & xArm-7 & 74.00\% & 54.00\% & 8.00\% & 94.00\% & 28.00\% & 50.00\% & 51.33\% & \\ 		
          & WidowX AI & 86.00\% & 60.00\% & 8.00\% & 48.00\% & 0.00\% & 68.00\% & 45.00\% & \\
         \hhline{-|---------}
         						
         \multicolumn{1}{l|}{\multirow{4}{*}{{UniAct}~\cite{VLA:UniACT}}} & Panda & 92.00\% & 88.00\% & 34.00\% & 92.00\% & 4.00\% & 0.00\% & 51.67\% & \multicolumn{1}{c}{\multirow{4}{*}{43.67\%}} \\ 
          & xArm-6 & 92.00\% & 94.00\% & 2.00\% & 94.00\% & 30.00\% & 4.00\% & 52.67\% &\\ 
          & xArm-7 & 94.00\% & 62.00\% & 12.00\% & 82.00\% & 14.00\% & 44.00\% & 51.33\% & \\ 
          & WidowX AI & 64.00\% & 38.00\% & 6.00\% & 4.00\% & 2.00\% & 0.00\% & 19.00\% & \\
         \hhline{-|---------}
         						
         \multicolumn{1}{l|}{\multirow{4}{*}{{GR00T-N1}~\cite{VLA:GR00T_N1}}} & Panda & 92.00\% & 100.00\% & 26.00\% & 58.00\% & 64.00\% & 66.00\% & 67.67\% & \multicolumn{1}{c}{\multirow{4}{*}{65.08\%}} \\ 
          & xArm-6 & 100.00\% & 100.00\% & 34.00\% & 100.00\% & 68.00\% & 14.00\% & 69.33\% &\\ 
          & xArm-7 &  94.00\% & 90.00\% & 14.00\% & 98.00\% & 56.00\% & 32.00\% & 64.00\% & \\ 
          & WidowX AI & 100.00\% & 52.00\% & 22.00\% & 60.00\% & 44.00\% & 78.00\% & 59.33\% & \\
         \hhline{-|---------}
         \multicolumn{1}{l|}{\multirow{4}{*}{{GR00T-N1.5}~\cite{GMLM:GR00T_N15}}} & Panda & 92.00\% & 100.00\% & 54.00\% & 100.00\% & 68.00\% & 68.00\% & 80.33\% & \multicolumn{1}{c}{\multirow{4}{*}{75.42\%}} \\ 
          & xArm-6 & 90.00\% & 88.00\% & 82.00\% & 100.00\% & 100.00\% & 48.00\% & 84.67\% &\\ 
          & xArm-7 & 98.00\% & 92.00\% & 50.00\% & 94.00\% & 80.00\% & 14.00\% & 71.33\% & \\ 
          & WidowX AI & 100.00\% & 70.00\% & 62.00\% & 48.00\% & 60.00\% & 52.00\% & 65.33\% & \\ 
        \hhline{-|---------}
         \multicolumn{10}{l}{\cellcolor{gray!20}\raisebox{0.2em}[0pt][0pt]{\downtriangle} \raisebox{-0.2em}[0pt][0pt]{\textbf{From-scratch Models}}} \\ [2pt] 
         \multicolumn{1}{l|}{\multirow{4}{*}{{Diffusion Policy}~\cite{VLA:DP}}} & Panda & 58.00\% & 100.00\% & 2.00\% & 92.00\% & 30.00\% & 28.00\% &  51.67\% & \multicolumn{1}{c}{\multirow{4}{*}{55.50\%}} \\ 
          & xArm-6 & 98.00\% & 96.00\% & 14.00\% & 86.00\% & 40.00\% & 50.00\% & 64.00\% &\\ 
          & xArm-7 & 90.00\% & 82.00\% & 18.00\% & 90.00\% & 72.00\% & 46.00\% & 66.33\% & \\ 
          & WidowX AI & 74.00\% & 76.00\% & 2.00\% & 16.00\% & 6.00\% & 66.00\% & 40.00\% & \\
          \hhline{-|---------}
          \multicolumn{1}{l|}{\multirow{4}{*}{Diffusion Policy\textsuperscript{$\dagger$}~\cite{VLA:DP}}} & Panda &  82.00\% & 100.00\% & 18.00\% & 100.00\% & 32.00\% & 64.00\% & 66.00\% & \multicolumn{1}{c}{\multirow{4}{*}{64.08\%}} \\ 
          & xArm-6 & 90.00\% & 96.00\% & 24.00\% & 94.00\% & 60.00\% & 50.00\% & 69.00\% &\\ 
          & xArm-7 &  96.00\% & 88.00\% & 18.00\% & 100.00\% & 66.00\% & 48.00\% & 69.33\% & \\ 
          & WidowX AI & 80.00\% & 84.00\% & 6.00\% & 58.00\% & 0.00\% & 84.00\% & 52.00\% & \\
          \hhline{-|---------}
          \multicolumn{1}{l|}{\multirow{4}{*}{Diffusion Policy\textsuperscript{$\star$}~\cite{VLA:DP}}} & Panda &  88.00\% & 98.00\% & 36.00\% & 100.00\% & 50.00\% & 72.00\% &  74.00\% & \multicolumn{1}{c}{\multirow{4}{*}{71.67\%}} \\ 
          & xArm-6 & 98.00\% & 98.00\% & 58.00\% & 100.00\% & 70.00\% & 52.00\% & 79.33\% &\\ 
          & xArm-7 & 98.00\% & 92.00\% & 48.00\% & 100.00\% & 84.00\% & 48.00\% & 78.33\% & \\ 
          & WidowX AI &  88.00\% & 88.00\% & 26.00\% & 52.00\% & 2.00\% & 74.00\% & 55.00\% & \\
         \hhline{-|---------}
         \multicolumn{1}{l|}{\multirow{4}{*}{{GR00T-N1.5}~\cite{GMLM:GR00T_N15}}} & Panda &  100.00\% & 98.00\% & 70.00\% & 82.00\% & 68.00\% & 78.00\% & 82.67\% & \multicolumn{1}{c}{\multirow{4}{*}{72.75\%}} \\ 
          & xArm-6 & 86.00\% & 98.00\% & 66.00\% & 98.00\% & 92.00\% & 6.00\% & 74.33\% &\\ 
          & xArm-7 & 94.00\% & 88.00\% & 42.00\% & 98.00\% & 86.00\% & 16.00\% & 70.67\% & \\ 
          & WidowX AI & 94.00\% & 82.00\% & 54.00\% & 74.00\% & 0.00\% & 76.00\% & 63.33\% & \\ 
         \hhline{-|---------}
         \multicolumn{1}{l|}{\multirow{4}{*}{\textbf{BLM$_1$}}} & Panda &  86.00\% &  100.00\% &  60.00\% &  100.00\% &  96.00\% &  66.00\% &  84.67\% & \multicolumn{1}{c}{\multirow{4}{*}{75.83\%}} \\ 
          & xArm-6 & 92.00\% & 100.00\% & 86.00\% & 100.00\% & 94.00\% & 10.00\% & 80.33\% &\\ 
          & xArm-7 & 72.00\% & 88.00\% & 56.00\% & 100.00\% & 78.00\% & 54.00\% & 74.67\% & \\ 
          & WidowX AI & 100.00\% & 68.00\% & 44.00\% & 84.00\% & 2.00\% & 84.00\% & 63.67\% & \\ 
        \Xhline{0.7pt}
    \end{tabular}
    \label{tab:physical_space_all}
    }
\end{table*}

%% file: Tables/appendix_hyperparameter.tex
\begin{table}[ht]
    \centering
    \captionsetup{belowskip=-5pt, aboveskip=5pt} 
    \caption{\textbf{Hyper-parameters used in the Stage I training of BLM$_1$.}}
     
    \setlength{\tabcolsep}{60.5pt}
    \footnotesize
    \scalebox{0.9}{ 
    \begin{tabular}{lc}
        \toprule
        Configurations & Values  \\
        \midrule
        Max token length & 4,096 \\
        Max pixel length & 451,584 \\
        Min pixel length & 12,544 \\
        Freeze vision tower & True \\
        Freeze multimodal projector & True\\
        Freeze language model & False\\
        Optimizer & AdamW \\
        Optimizer hyperparameters & $\beta_1$=0.9, $\beta_2$=0.999, $\epsilon$=1e-8\\
        Weight decay &  0.01   \\
        Peak learning rate    &  1e-6   \\
        Learning rate schedule & cosine decay \\
        Training epochs & 1 \\
        Training steps & 18,046 \\
        Warm-up ratio & 0.01 \\
        Global batch size & 128\\
        Gradient accumulation & 1\\
        Numerical precision & bfloat16\\
        \bottomrule
    \end{tabular}}
    \vspace{0.5em}
    \label{tab:training_setup}
    \vspace{-2em}
\end{table}

%% file: Tables/appendix_hyperparameter_2.tex
\begin{table}[ht]
    \centering
    \captionsetup{belowskip=-5pt, aboveskip=5pt} 
    \caption{\textbf{Hyper-parameters used in Stage II training of BLM$_1$.}}
     
    \setlength{\tabcolsep}{57pt}
    \footnotesize
    \scalebox{0.9}{ 
    \begin{tabular}{lc}
        \toprule
        Configurations & Values  \\
        \midrule
        Optimizer & AdamW \\
        Optimizer hyper-parameters & $\beta_1$=0.9, $\beta_2$=0.999, $\epsilon$=1e-8\\
        Weight decay &  0.01   \\
        Peak learning rate    &  1e-4   \\
        Learning rate schedule & cosine decay \\
        Training steps & 200,000 \\
        Warm-up ratio & 0.03 \\
        Global batch size & 128\\
        Gradient accumulation & 1\\
        Numerical precision & float32\\
        Feature layer select & 14\\
        Action horizon & 16\\
        Max state dimension & 64\\
        Max action dimension & 32\\
        DiT hidden dimension & 1,536\\
        DiT layer number & 16\\
        DiT attention head dimension & 48\\
        DiT attention head number & 32\\
        DiT attention dropout & 0.2\\
        Noise-beta-distribution & $\alpha$=1.5, $\beta$=1.0\\
        Noise schedule & 0.999\\
        Inference timesteps & 4\\
        Timesteps buckets & 1,000\\
        Vl-self-attention layers number & 4\\
        Vl-self-attention head dimension & 48\\
        Vl-self-attention head number & 32\\
        Vl-self-attention dropout & 0.2\\
        \bottomrule
    \end{tabular}}
    \vspace{0.5em}
    \label{tab:Stage_II_training_setup}
    \vspace{-2em}
\end{table}

%% file: Tables/hyper-parameters_stage2_baselines.tex
\begin{table}[ht]
    \centering
    \captionsetup{belowskip=-5pt, aboveskip=5pt} 
    \caption{\textbf{Hyper-parameters used in the Stage II training of the robotic baselines.}}
    \label{tab:baseline_setup}
    \setlength{\tabcolsep}{30.2pt}
    \renewcommand{\arraystretch}{1}
    \footnotesize
    \scalebox{0.9}{ 
    \begin{tabular}{l|cc}
        \Xhline{0.7pt}
        Models & Configurations & Values  \\
        \hline
        \multicolumn{3}{l}{\cellcolor{gray!20} \raisebox{0.2em}[0pt][0pt]{\downtriangle} \raisebox{-0.2em}[0pt][0pt]{\textbf{Pre-trained Models}}} \\ [2pt] 
        \multirow{8}{*}{
        \makecell[l]{
            {\Large $\pi$}\textsubscript{0}~\cite{VLA:Pi0} \\ \\
            {\Large $\pi$}\textsubscript{0}\textsuperscript{$\dagger$}~\cite{VLA:Pi0}}
        } & Batch size & 32 \\
        & Max training steps & 100,000 \\
        & Peak learning rate & 2e-5 \\
        & Decay learning rate & 1.5e-6 \\
        & Decay step & 100,000 \\
        & Weight decay &  1e-5 \\
        & Warm-up ratio &  0.05 \\
        & Learning rate schedule & cosine decay \\
        \hhline{-|--}
        \multirow{5}{*}{{HPT}~\cite{VLA:HPT}} & Batch size & 768 \\
        & Epoch & 1,000 \\
        & Warm-up learning rate & 1e-10 \\
        & Warm-up step &  2,000 \\
        & Learning rate & 1e-5 \\
        \hhline{-|--}
        \multirow{3}{*}{{UniAct}~\cite{VLA:UniACT}} & Batch size & 64 \\
        & Peak learning rate & 1e-4 \\
        & Max training steps & 1,500,000 \\
        & Action chunk & 8 \\
        \hhline{-|--}
        \multirow{6}{*}{{GR00T-N1}~\cite{VLA:GR00T_N1}} & Batch size & 32 \\
        & Max training steps & 300,000 \\
        & Peak learning rate & 1e-4 \\
        & Weight decay &  1e-5 \\
        & Warm-up ratio &  0.05 \\
        & Learning rate schedule & cosine decay \\
        \hhline{-|--}
        \multirow{6}{*}{{GR00T-N1.5}~\cite{GMLM:GR00T_N15}} & Batch size & 128 \\
        & Max training steps & 300,000 \\
        & Peak learning rate & 1e-4 \\
        & Weight decay &  1e-5 \\
        & Warm-up ratio &  0.05 \\
        & Learning rate schedule & cosine decay \\
        \hhline{-|--}
        \multicolumn{3}{l}{ \cellcolor{gray!20} \raisebox{0.2em}[0pt][0pt]{\downtriangle} \raisebox{-0.2em}[0pt][0pt]
        {\textbf{From-scratch Models}}} \\ [2pt] 
        \multirow{6}{*}{
        \makecell[l]{
            {Diffusion Policy}~\cite{VLA:DP} \\ \\
            {Diffusion Policy}\textsuperscript{$\dagger$}~\cite{VLA:DP} \\ \\
            {Diffusion Policy}\textsuperscript{$\star$}~\cite{VLA:DP}
        }
        } & Batch size & 64 \\
        & Warm-up steps & 1,000 \\
        & Learning rate & 3e-4 \\
        & Transformer weight decay &  1e-3 \\
        & Observation encoder weight decay &  1e-6 \\
        & Crop shape & (243,243) \\
        \hhline{-|--}
        \multirow{6}{*}{\makecell[l]{GR00T-N1.5}~\cite{GMLM:GR00T_N15}} & Batch size & 32 \\
        & Max training steps & 300,000 \\
        & Peak learning rate & 1e-4 \\
        & Weight decay &  1e-5 \\
        & Warm-up ratio &  0.05 \\
        & Learning rate schedule & cosine decay \\
        \Xhline{0.7pt}
    \end{tabular}}
\end{table}

%% file: template.bbl
\begin{thebibliography}{10}

\bibitem{MLLM:Survey}
Shukang Yin, Chaoyou Fu, Sirui Zhao, Ke~Li, Xing Sun, Tong Xu, and Enhong Chen.
\newblock A survey on multimodal large language models.
\newblock {\em arXiv preprint arXiv:2306.13549}, 2024.

\bibitem{MLLM:Qwen2.5-VL}
Shuai Bai, Keqin Chen, Xuejing Liu, Jialin Wang, Wenbin Ge, Sibo Song, Kai Dang, Peng Wang, Shijie Wang, Jun Tang, et~al.
\newblock Qwen2.5-vl technical report.
\newblock {\em arXiv preprint arXiv:2502.13923}, 2025.

\bibitem{MLLM:GPT-4}
Josh Achiam, Steven Adler, Sandhini Agarwal, Lama Ahmad, Ilge Akkaya, Florencia~Leoni Aleman, Diogo Almeida, Janko Altenschmidt, Sam Altman, Shyamal Anadkat, et~al.
\newblock Gpt-4 technical report.
\newblock {\em arXiv preprint arXiv:2303.08774}, 2023.

\bibitem{MLLM:MLLM-Driving}
Can Cui, Yunsheng Ma, Xu~Cao, Wenqian Ye, Yang Zhou, Kaizhao Liang, Jintai Chen, Juanwu Lu, Zichong Yang, Kuei-Da Liao, et~al.
\newblock A survey on multimodal large language models for autonomous driving.
\newblock In {\em Proceedings of the IEEE/CVF Winter Conference on Applications of Computer Vision}, pages 958--979, 2024.

\bibitem{MLLM:AI-assistant}
Chunyuan Li, Zhe Gan, Zhengyuan Yang, Jianwei Yang, Linjie Li, Lijuan Wang, and Jianfeng Gao.
\newblock Multimodal foundation models: From specialists to general-purpose assistants.
\newblock {\em arXiv preprint arXiv:2309.10020}, 2023.

\bibitem{MLLM:MLLM-Robot}
Jiaqi Wang, Zihao Wu, Yiwei Li, Hanqi Jiang, Peng Shu, Enze Shi, Huawen Hu, Chong Ma, Yiheng Liu, Xuhui Wang, et~al.
\newblock Large language models for robotics: Opportunities, challenges, and perspectives.
\newblock {\em arXiv preprint arXiv:2401.04334}, 2024.

\bibitem{VLA:RT-2}
Brianna Zitkovich, Tianhe Yu, Sichun Xu, Peng Xu, Ted Xiao, Fei Xia, Jialin Wu, Paul Wohlhart, Stefan Welker, Ayzaan Wahid, et~al.
\newblock Rt-2: Vision-language-action models transfer web knowledge to robotic control.
\newblock In {\em Proceedings of the Conference on Robot Learning}, pages 2165--2183, 2023.

\bibitem{VLA:OpenVLA}
Moo~Jin Kim, Karl Pertsch, Siddharth Karamcheti, Ted Xiao, Ashwin Balakrishna, Suraj Nair, Rafael Rafailov, Ethan Foster, Grace Lam, Pannag Sanketi, et~al.
\newblock Openvla: An open-source vision-language-action model.
\newblock {\em arXiv preprint arXiv:2406.09246}, 2024.

\bibitem{VLA:Pi0.5}
Physical Intelligence, Kevin Black, Noah Brown, James Darpinian, Karan Dhabalia, Danny Driess, Adnan Esmail, Michael Equi, Chelsea Finn, Niccolo Fusai, et~al.
\newblock {$\pi_{0.5}$: A Vision–Language–Action Model with Open-World Generalization}.
\newblock {\em arXiv preprint arXiv:2504.16054}, 2025.

\bibitem{ELLM:Embodiedgpt}
Yao Mu, Qinglong Zhang, Mengkang Hu, Wenhai Wang, Mingyu Ding, Jun Jin, Bin Wang, Jifeng Dai, Yu~Qiao, and Ping Luo.
\newblock Embodiedgpt: Vision-language pre-training via embodied chain of thought.
\newblock In {\em Advances in Neural Information Processing Systems}, pages 25081--25094, 2023.

\bibitem{ELLM:Cosmos-Reason1}
Alisson Azzolini, Junjie Bai, Hannah Brandon, Jiaxin Cao, Prithvijit Chattopadhyay, Huayu Chen, Jinju Chu, Yin Cui, Jenna Diamond, Yifan Ding, et~al.
\newblock Cosmos-reason1: From physical common sense to embodied reasoning.
\newblock {\em arXiv preprint arXiv:2503.15558}, 2025.

\bibitem{VLA:FAST}
Karl Pertsch, Kyle Stachowicz, Brian Ichter, Danny Driess, Suraj Nair, Quan Vuong, Oier Mees, Chelsea Finn, and Sergey Levine.
\newblock Fast: Efficient action tokenization for vision-language-action models.
\newblock {\em arXiv preprint arXiv:2501.09747}, 2025.

\bibitem{VLA:DP}
Cheng Chi, Siyuan Feng, Yilun Du, Zhenjia Xu, Eric Cousineau, Benjamin Burchfiel, and Shuran Song.
\newblock Diffusion policy: Visuomotor policy learning via action diffusion.
\newblock In {\em Proceedings of Robotics: Science and Systems}, 2023.

\bibitem{VLA:Pi0}
Kevin Black, Noah Brown, Danny Driess, Adnan Esmail, Michael Equi, Chelsea Finn, Niccolo Fusai, Lachy Groom, Karol Hausman, Brian Ichter, et~al.
\newblock $\pi$0: A vision-language-action flow model for general robot control.
\newblock {\em arXiv preprint arXiv:2410.24164}, 2024.

\bibitem{VLA:RDT}
Songming Liu, Lingxuan Wu, Bangguo Li, Hengkai Tan, Huayu Chen, Zhengyi Wang, Ke~Xu, Hang Su, and Jun Zhu.
\newblock Rdt-1b: a diffusion foundation model for bimanual manipulation.
\newblock {\em arXiv preprint arXiv:2410.07864}, 2024.

\bibitem{VLA:DiVLA}
Junjie Wen, Minjie Zhu, Yichen Zhu, Zhibin Tang, Jinming Li, Zhongyi Zhou, Chengmeng Li, Xiaoyu Liu, Yaxin Peng, Chaomin Shen, et~al.
\newblock Diffusion-vla: Generalizable and interpretable robot foundation model via self-generated reasoning.
\newblock {\em arXiv preprint arXiv:2412.03293}, 2024.

\bibitem{GMLM:ChatVLA-2}
Zhongyi Zhou, Yichen Zhu, Junjie Wen, Chaomin Shen, and Yi~Xu.
\newblock Chatvla-2: Vision-language-action model with open-world embodied reasoning from pretrained knowledge.
\newblock {\em arXiv preprint arXiv:2505.21906}, 2025.

\bibitem{GMLM:Robomamba}
Jiaming Liu, Mengzhen Liu, Zhenyu Wang, Pengju An, Xiaoqi Li, Kaichen Zhou, Senqiao Yang, Renrui Zhang, Yandong Guo, and Shanghang Zhang.
\newblock Robomamba: Efficient vision-language-action model for robotic reasoning and manipulation.
\newblock In {\em Advances in Neural Information Processing Systems}, pages 40085--40110, 2024.

\bibitem{GMLM:Gemini-Robotics}
Gemini~Robotics Team, Saminda Abeyruwan, Joshua Ainslie, Jean-Baptiste Alayrac, Montserrat~Gonzalez Arenas, Travis Armstrong, Ashwin Balakrishna, Robert Baruch, Maria Bauza, Michiel Blokzijl, et~al.
\newblock Gemini robotics: Bringing ai into the physical world.
\newblock {\em arXiv preprint arXiv:2503.20020}, 2025.

\bibitem{GMLM:GR00T_N15}
Johan Bjorck, Valts Blukis, Fernando Castañeda, Nikita Cherniadev, Xingye Da, Runyu Ding, Linxi Fan, Yu~Fang, Dieter Fox, Fengyuan Hu, Spencer Huang, Joel Jang, Xiaowei Jiang, Kaushil Kundalia, Jan Kautz, Zhiqi Li, Kevin Lin, Zongyu Lin, Loic Magne, Yunze Man, Ajay Mandlekar, Avnish Narayan, Soroush Nasiriany, Scott Reed, You~Liang Tan, Guanzhi Wang, Jing Wang, Qi~Wang, Shihao Wang, Jiannan Xiang, Yuqi Xie, Yinzhen Xu, Seonghyeon Ye, Zhiding Yu, Yizhou Zhao, Zhe Zhang, Ruijie Zheng, and Yuke Zhu.
\newblock Gr00t n1.5: An improved open foundation model for generalist humanoid robots.
\newblock {\em NVIDIA Research Blog:2506.11}, 2025.

\bibitem{ManiSkill3}
Stone Tao, Fanbo Xiang, Arth Shukla, Yuzhe Qin, Xander Hinrichsen, Xiaodi Yuan, Chen Bao, Xinsong Lin, Yulin Liu, Tse-kai Chan, et~al.
\newblock Maniskill3: Gpu parallelized robotics simulation and rendering for generalizable embodied ai.
\newblock {\em arXiv preprint arXiv:2410.00425}, 2024.

\bibitem{LLM:DeepSeek-R1}
Daya Guo, Dejian Yang, Haowei Zhang, Junxiao Song, Ruoyu Zhang, Runxin Xu, Qihao Zhu, Shirong Ma, Peiyi Wang, Xiao Bi, et~al.
\newblock Deepseek-r1: Incentivizing reasoning capability in llms via reinforcement learning.
\newblock {\em arXiv preprint arXiv:2501.12948}, 2025.

\bibitem{LLM:LLaMA}
Hugo Touvron, Thibaut Lavril, Gautier Izacard, Xavier Martinet, Marie-Anne Lachaux, Timoth{\'e}e Lacroix, Baptiste Rozi{\`e}re, Naman Goyal, Eric Hambro, Faisal Azhar, Aur{\'e}lien Rodriguez, Armand Joulin, Edouard Grave, and Guillaume Lample.
\newblock Llama: Open and efficient foundation language models.
\newblock {\em arXiv preprint arXiv:2302.13971}, 2023.

\bibitem{LLM:Mixtral}
Albert~Q Jiang, Alexandre Sablayrolles, Antoine Roux, Arthur Mensch, Blanche Savary, Chris Bamford, Devendra~Singh Chaplot, Diego de~las Casas, Emma~Bou Hanna, Florian Bressand, et~al.
\newblock Mixtral of experts.
\newblock {\em arXiv preprint arXiv:2401.04088}, 2024.

\bibitem{LLM:Phi-4}
Marah Abdin, Jyoti Aneja, Harkirat Behl, S{\'e}bastien Bubeck, Ronen Eldan, Suriya Gunasekar, Michael Harrison, Russell~J Hewett, Mojan Javaheripi, Piero Kauffmann, et~al.
\newblock Phi-4 technical report.
\newblock {\em arXiv preprint arXiv:2412.08905}, 2024.

\bibitem{MLLM:LLaVA}
Haotian Liu, Chunyuan Li, Qingyang Wu, and Yong~Jae Lee.
\newblock Visual instruction tuning.
\newblock In {\em Advances in Neural Information Processing Systems}, pages 34892--34916, 2023.

\bibitem{MLLM:BLIP-2}
Junnan Li, Dongxu Li, Silvio Savarese, and Steven Hoi.
\newblock Blip-2: Bootstrapping language-image pre-training with frozen image encoders and large language models.
\newblock In {\em Proceedings of the International Conference on Machine Learning}, pages 19730--19742, 2023.

\bibitem{VFM:CLIP}
Alec Radford, Jong~Wook Kim, Chris Hallacy, Aditya Ramesh, Gabriel Goh, Sandhini Agarwal, Girish Sastry, Amanda Askell, Pamela Mishkin, Jack Clark, et~al.
\newblock Learning transferable visual models from natural language supervision.
\newblock In {\em Proceedings of the International Conference on Machine Learning}, pages 8748--8763, 2021.

\bibitem{VFM:ALIGN}
Chao Jia, Yinfei Yang, Ye~Xia, Yi-Ting Chen, Zarana Parekh, Hieu Pham, Quoc~V. Le, Yun-Hsuan Sung, Zhen Li, and Tom Duerig.
\newblock Scaling up visual and vision-language representation learning with noisy text supervision.
\newblock In {\em Proceedings of the International Conference on Machine Learning}, pages 4904--4916, 2021.

\bibitem{VFM:EVA}
Yuxin Fang, Wen Wang, Binhui Xie, Quan-Sen Sun, Ledell~Yu Wu, Xinggang Wang, Tiejun Huang, Xinlong Wang, and Yue Cao.
\newblock Eva: Exploring the limits of masked visual representation learning at scale.
\newblock In {\em Proceedings of the IEEE/CVF Conference on Computer Vision and Pattern Recognition}, pages 19358--19369, 2022.

\bibitem{VFM:ViT}
Alexey Dosovitskiy, Lucas Beyer, Alexander Kolesnikov, Dirk Weissenborn, Xiaohua Zhai, Thomas Unterthiner, Mostafa Dehghani, Matthias Minderer, Georg Heigold, Sylvain Gelly, Jakob Uszkoreit, and Neil Houlsby.
\newblock An image is worth 16x16 words: Transformers for image recognition at scale.
\newblock In {\em Proceedings of the International Conference on Learning Representations}, 2021.

\bibitem{VFM:MAE}
Kaiming He, Xinlei Chen, Saining Xie, Yanghao Li, Piotr Doll{\'{a}}r, and Ross~B. Girshick.
\newblock Masked autoencoders are scalable vision learners.
\newblock In {\em Proceedings of the IEEE/CVF Conference on Computer Vision and Pattern Recognition}, pages 15979--15988, 2022.

\bibitem{VFM:SigLIP}
Xiaohua Zhai, Basil Mustafa, Alexander Kolesnikov, and Lucas Beyer.
\newblock Sigmoid loss for language image pre-training.
\newblock In {\em Proceedings of the IEEE/CVF International Conference on Computer Vision}, pages 11975--11986, 2023.

\bibitem{RLHF1}
Long Ouyang, Jeffrey Wu, Xu~Jiang, Diogo Almeida, Carroll Wainwright, Pamela Mishkin, Chong Zhang, Sandhini Agarwal, Katarina Slama, Alex Ray, et~al.
\newblock Training language models to follow instructions with human feedback.
\newblock In {\em Advances in Neural Information Processing Systems}, pages 27730--27744, 2022.

\bibitem{DPO}
Rafael Rafailov, Archit Sharma, Eric Mitchell, Christopher~D Manning, Stefano Ermon, and Chelsea Finn.
\newblock Direct preference optimization: Your language model is secretly a reward model.
\newblock In {\em Advances in Neural Information Processing Systems}, pages 53728--53741, 2023.

\bibitem{Datasets:GQA}
Drew~A. Hudson and Christopher~D. Manning.
\newblock Gqa: A new dataset for real-world visual reasoning and compositional question answering.
\newblock In {\em Proceedings of the IEEE/CVF Conference on Computer Vision and Pattern Recognition}, pages 6700--6709, 2019.

\bibitem{Datasets:MMBench}
Yuan Liu, Haodong Duan, Yuanhan Zhang, Bo~Li, Songyang Zhang, Wangbo Zhao, Yike Yuan, Jiaqi Wang, Conghui He, Ziwei Liu, Kai Chen, and Dahua Lin.
\newblock Mmbench: Is your multi-modal model an all-around player?
\newblock {\em arXiv preprint arXiv:2307.06281}, 2023.

\bibitem{Datasets:MMMU}
Xiang Yue, Yuansheng Ni, Kai Zhang, Tianyu Zheng, Ruoqi Liu, Ge~Zhang, Samuel Stevens, Dongfu Jiang, Weiming Ren, Yuxuan Sun, et~al.
\newblock Mmmu: A massive multi-discipline multimodal understanding and reasoning benchmark for expert agi.
\newblock {\em arXiv preprint arXiv:2311.16502}, 2023.

\bibitem{Datasets:TextVQA}
Amanpreet Singh, Vivek Natarajan, Meet Shah, Yu~Jiang, Xinlei Chen, Dhruv Batra, Devi Parikh, and Marcus Rohrbach.
\newblock Towards vqa models that can read.
\newblock In {\em Proceedings of the IEEE/CVF Conference on Computer Vision and Pattern Recognition}, pages 8317--8326, 2019.

\bibitem{MLLM:Flamingo}
Jean-Baptiste Alayrac, Jeff Donahue, Pauline Luc, Antoine Miech, Iain Barr, Yana Hasson, Karel Lenc, Arthur Mensch, Katherine Millican, Malcolm Reynolds, et~al.
\newblock Flamingo: a visual language model for few-shot learning.
\newblock In {\em Advances in Neural Information Processing Systems}, pages 23716--23736, 2022.

\bibitem{MLLM:Gemini}
Gemini Team, Rohan Anil, Sebastian Borgeaud, Jean-Baptiste Alayrac, Jiahui Yu, Radu Soricut, Johan Schalkwyk, Andrew~M Dai, Anja Hauth, Katie Millican, et~al.
\newblock Gemini: a family of highly capable multimodal models.
\newblock {\em arXiv preprint arXiv:2312.11805}, 2023.

\bibitem{MLLM:InternVL}
Zhe Chen, Jiannan Wu, Wenhai Wang, Weijie Su, Guo Chen, Sen Xing, Muyan Zhong, Qinglong Zhang, Xizhou Zhu, Lewei Lu, et~al.
\newblock Internvl: Scaling up vision foundation models and aligning for generic visual-linguistic tasks.
\newblock In {\em Proceedings of the IEEE/CVF Conference on Computer Vision and Pattern Recognition}, pages 24185--24198, 2024.

\bibitem{ELLM:Robopoint}
Wentao Yuan, Jiafei Duan, Valts Blukis, Wilbert Pumacay, Ranjay Krishna, Adithyavairavan Murali, Arsalan Mousavian, and Dieter Fox.
\newblock Robopoint: A vision-language model for spatial affordance prediction for robotics.
\newblock {\em arXiv preprint arXiv:2406.10721}, 2024.

\bibitem{ELLM:Palm-e}
Danny Driess, Fei Xia, Mehdi~SM Sajjadi, Corey Lynch, Aakanksha Chowdhery, Brian Ichter, Ayzaan Wahid, Jonathan Tompson, Quan Vuong, Tianhe Yu, et~al.
\newblock Palm-e: An embodied multimodal language model.
\newblock {\em arXiv preprint arXiv:2303.03378}, 2023.

\bibitem{ELLM:3D-LLM}
Yining Hong, Haoyu Zhen, Peihao Chen, Shuhong Zheng, Yilun Du, Zhenfang Chen, and Chuang Gan.
\newblock 3d-llm: Injecting the 3d world into large language models.
\newblock In {\em Advances in Neural Information Processing Systems}, pages 20482--20494, 2023.

\bibitem{Inner_monologue}
Wenlong Huang, Fei Xia, Ted Xiao, Harris Chan, Jacky Liang, Pete Florence, Andy Zeng, Jonathan Tompson, Igor Mordatch, Yevgen Chebotar, et~al.
\newblock Inner monologue: Embodied reasoning through planning with language models.
\newblock {\em arXiv preprint arXiv:2207.05608}, 2022.

\bibitem{CodeAP}
Jacky Liang, Wenlong Huang, Fei Xia, Peng Xu, Karol Hausman, Brian Ichter, Peter~R. Florence, and Andy Zeng.
\newblock Code as policies: Language model programs for embodied control.
\newblock In {\em Proceedings of the IEEE International Conference on Robotics and Automation}, pages 9493--9500, 2022.

\bibitem{SayCan}
Anthony Brohan, Yevgen Chebotar, Chelsea Finn, Karol Hausman, Alexander Herzog, Daniel Ho, Julian Ibarz, Alex Irpan, Eric Jang, Ryan Julian, et~al.
\newblock Do as i can, not as i say: Grounding language in robotic affordances.
\newblock In {\em Proceedings of the Conference on Robot Learning}, pages 287--318, 2023.

\bibitem{ELLM:Point-LLM}
Runsen Xu, Xiaolong Wang, Tai Wang, Yilun Chen, Jiangmiao Pang, and Dahua Lin.
\newblock Pointllm: Empowering large language models to understand point clouds.
\newblock In {\em Proceedings of the European Conference on Computer Vision}, pages 131--147, 2024.

\bibitem{ELLM:LEO}
Jiangyong Huang, Silong Yong, Xiaojian Ma, Xiongkun Linghu, Puhao Li, Yan Wang, Qing Li, Song-Chun Zhu, Baoxiong Jia, and Siyuan Huang.
\newblock An embodied generalist agent in 3d world.
\newblock {\em arXiv preprint arXiv:2311.12871}, 2023.

\bibitem{ELLM:Magma}
Jianwei Yang, Reuben Tan, Qianhui Wu, Ruijie Zheng, Baolin Peng, Yongyuan Liang, Yu~Gu, Mu~Cai, Seonghyeon Ye, Joel Jang, et~al.
\newblock Magma: A foundation model for multimodal ai agents.
\newblock In {\em Proceedings of the IEEE/CVF Conference on Computer Vision and Pattern Recognition}, pages 14203--14214, 2025.

\bibitem{ELLM:RoboBrain}
Yuheng Ji, Huajie Tan, Jiayu Shi, Xiaoshuai Hao, Yuan Zhang, Hengyuan Zhang, Pengwei Wang, Mengdi Zhao, Yao Mu, Pengju An, et~al.
\newblock Robobrain: A unified brain model for robotic manipulation from abstract to concrete.
\newblock In {\em Proceedings of the IEEE/CVF Conference on Computer Vision and Pattern Recognition}, pages 1724--1734, 2025.

\bibitem{ELLM:RoboBrain2}
BAAI~RoboBrain Team, Mingyu Cao, Huajie Tan, Yuheng Ji, Minglan Lin, Zhiyu Li, Zhou Cao, Pengwei Wang, Enshen Zhou, Yi~Han, et~al.
\newblock Robobrain 2.0 technical report.
\newblock {\em arXiv preprint arXiv:2507.02029}, 2025.

\bibitem{ELLM:VeBrain}
Gen Luo, Ganlin Yang, Ziyang Gong, Guanzhou Chen, Haonan Duan, Erfei Cui, Ronglei Tong, Zhi Hou, Tianyi Zhang, Zhe Chen, et~al.
\newblock Visual embodied brain: Let multimodal large language models see, think, and control in spaces.
\newblock {\em arXiv preprint arXiv:2506.00123}, 2025.

\bibitem{CoT}
Jason Wei, Xuezhi Wang, Dale Schuurmans, Maarten Bosma, Fei Xia, Ed~Chi, Quoc~V Le, Denny Zhou, et~al.
\newblock Chain-of-thought prompting elicits reasoning in large language models.
\newblock In {\em Advances in Neural Information Processing Systems}, pages 24824--24837, 2022.

\bibitem{CoT1}
Micha{\l} Zawalski, William Chen, Karl Pertsch, Oier Mees, Chelsea Finn, and Sergey Levine.
\newblock Robotic control via embodied chain-of-thought reasoning.
\newblock {\em arXiv preprint arXiv:2407.08693}, 2024.

\bibitem{VLA:GR-2}
Chi-Lam Cheang, Guangzeng Chen, Ya~Jing, Tao Kong, Hang Li, Yifeng Li, Yuxiao Liu, Hongtao Wu, Jiafeng Xu, Yichu Yang, et~al.
\newblock Gr-2: A generative video-language-action model with web-scale knowledge for robot manipulation.
\newblock {\em arXiv preprint arXiv:2410.06158}, 2024.

\bibitem{VLA:Octo}
Octo~Model Team, Dibya Ghosh, Homer Walke, Karl Pertsch, Kevin Black, Oier Mees, Sudeep Dasari, Joey Hejna, Tobias Kreiman, Charles Xu, et~al.
\newblock Octo: An open-source generalist robot policy.
\newblock {\em arXiv preprint arXiv:2405.12213}, 2024.

\bibitem{VLA:RT-1}
Anthony Brohan, Noah Brown, Justice Carbajal, Yevgen Chebotar, Joseph Dabis, Chelsea Finn, Keerthana Gopalakrishnan, Karol Hausman, Alex Herzog, Jasmine Hsu, et~al.
\newblock Rt-1: Robotics transformer for real-world control at scale.
\newblock {\em arXiv preprint arXiv:2212.06817}, 2022.

\bibitem{VLA:dita}
Zhi Hou, Tianyi Zhang, Yuwen Xiong, Haonan Duan, Hengjun Pu, Ronglei Tong, Chengyang Zhao, Xizhou Zhu, Yu~Qiao, Jifeng Dai, et~al.
\newblock Dita: Scaling diffusion transformer for generalist vision-language-action policy.
\newblock {\em arXiv preprint arXiv:2503.19757}, 2025.

\bibitem{VLA:HPT}
Lirui Wang, Xinlei Chen, Jialiang Zhao, and Kaiming He.
\newblock Scaling proprioceptive-visual learning with heterogeneous pre-trained transformers.
\newblock In {\em Advances in Neural Information Processing Systems}, pages 124420--124450, 2024.

\bibitem{VLA:UniACT}
Jinliang Zheng, Jianxiong Li, Dongxiu Liu, Yinan Zheng, Zhihao Wang, Zhonghong Ou, Yu~Liu, Jingjing Liu, Ya-Qin Zhang, and Xianyuan Zhan.
\newblock Universal actions for enhanced embodied foundation models.
\newblock In {\em Proceedings of the IEEE/CVF Conference on Computer Vision and Pattern Recognition}, pages 22508--22519, 2025.

\bibitem{VLA:GR00T_N1}
Johan Bjorck, Fernando Casta{\~n}eda, Nikita Cherniadev, Xingye Da, Runyu Ding, Linxi Fan, Yu~Fang, Dieter Fox, Fengyuan Hu, Spencer Huang, et~al.
\newblock Gr00t n1: An open foundation model for generalist humanoid robots.
\newblock {\em arXiv preprint arXiv:2503.14734}, 2025.

\bibitem{GMLM:ChatVLA}
Zhongyi Zhou, Yichen Zhu, Minjie Zhu, Junjie Wen, Ning Liu, Zhiyuan Xu, Weibin Meng, Ran Cheng, Yaxin Peng, Chaomin Shen, et~al.
\newblock Chatvla: Unified multimodal understanding and robot control with vision-language-action model.
\newblock {\em arXiv preprint arXiv:2502.14420}, 2025.

\bibitem{MLLM:PaLI}
Xi~Chen, Xiao Wang, Soravit Changpinyo, Anthony~J Piergiovanni, Piotr Padlewski, Daniel Salz, Sebastian Goodman, Adam Grycner, Basil Mustafa, Lucas Beyer, et~al.
\newblock Pali: A jointly-scaled multilingual language-image model.
\newblock {\em arXiv preprint arXiv:2209.06794}, 2022.

\bibitem{MoE1}
Damai Dai, Chengqi Deng, Chenggang Zhao, Rx~Xu, Huazuo Gao, Deli Chen, Jiashi Li, Wangding Zeng, Xingkai Yu, Y~Wu, et~al.
\newblock Deepseekmoe: Towards ultimate expert specialization in mixture-of-experts language models.
\newblock In {\em Proceedings of the Annual Meeting of the Association for Computational Linguistics}, pages 1280--1297, 2024.

\bibitem{MoE2}
Bin Lin, Zhenyu Tang, Yang Ye, Jiaxi Cui, Bin Zhu, Peng Jin, Jinfa Huang, Junwu Zhang, Yatian Pang, Munan Ning, et~al.
\newblock Moe-llava: Mixture of experts for large vision-language models.
\newblock {\em arXiv preprint arXiv:2401.15947}, 2024.

\bibitem{perceiver}
Andrew Jaegle, Felix Gimeno, Andy Brock, Oriol Vinyals, Andrew Zisserman, and Joao Carreira.
\newblock Perceiver: General perception with iterative attention.
\newblock In {\em Proceedings of the International Conference on Machine Learning}, pages 4651--4664, 2021.

\bibitem{FlowMatching}
Yaron Lipman, Ricky~TQ Chen, Heli Ben-Hamu, Maximilian Nickel, and Matt Le.
\newblock Flow matching for generative modeling.
\newblock In {\em Proceedings of the International Conference on Learning Representations}, 2023.

\bibitem{flare}
Ruijie Zheng, Jing Wang, Scott Reed, Johan Bjorck, Yu~Fang, Fengyuan Hu, Joel Jang, Kaushil Kundalia, Zongyu Lin, Loic Magne, et~al.
\newblock Flare: Robot learning with implicit world modeling.
\newblock {\em arXiv preprint arXiv:2505.15659}, 2025.

\bibitem{robovqa}
Pierre Sermanet, Tianli Ding, Jeffrey Zhao, Fei Xia, Debidatta Dwibedi, Keerthana Gopalakrishnan, Christine Chan, Gabriel Dulac-Arnold, Sharath Maddineni, Nikhil~J Joshi, et~al.
\newblock Robovqa: Multimodal long-horizon reasoning for robotics.
\newblock In {\em Proceedings of the IEEE International Conference on Robotics and Automation}, pages 645--652, 2024.

\bibitem{agibot}
Qingwen Bu, Jisong Cai, Li~Chen, Xiuqi Cui, Yan Ding, Siyuan Feng, Shenyuan Gao, Xindong He, Xuan Hu, Xu~Huang, et~al.
\newblock Agibot world colosseo: A large-scale manipulation platform for scalable and intelligent embodied systems.
\newblock {\em arXiv preprint arXiv:2503.06669}, 2025.

\bibitem{holoassist}
Xin Wang, Taein Kwon, Mahdi Rad, Bowen Pan, Ishani Chakraborty, Sean Andrist, Dan Bohus, Ashley Feniello, Bugra Tekin, Felipe~Vieira Frujeri, et~al.
\newblock Holoassist: An egocentric human interaction dataset for interactive ai assistants in the real world.
\newblock In {\em Proceedings of the IEEE/CVF International Conference on Computer Vision}, pages 20270--20281, 2023.

\bibitem{bridge}
Homer~Rich Walke, Kevin Black, Tony~Z Zhao, Quan Vuong, Chongyi Zheng, Philippe Hansen-Estruch, Andre~Wang He, Vivek Myers, Moo~Jin Kim, Max Du, et~al.
\newblock Bridgedata v2: A dataset for robot learning at scale.
\newblock In {\em Proceedings of the Conference on Robot Learning}, pages 1723--1736, 2023.

\bibitem{egoplan}
Yi~Chen, Yuying Ge, Yixiao Ge, Mingyu Ding, Bohao Li, Rui Wang, Ruifeng Xu, Ying Shan, and Xihui Liu.
\newblock Egoplan-bench: Benchmarking egocentric embodied planning with multimodal large language models.
\newblock {\em CoRR}, 2023.

\bibitem{EPIC-Kitchens}
Dima Damen, Hazel Doughty, Giovanni~Maria Farinella, Antonino Furnari, Evangelos Kazakos, Jian Ma, Davide Moltisanti, Jonathan Munro, Toby Perrett, Will Price, et~al.
\newblock Rescaling egocentric vision: Collection, pipeline and challenges for epic-kitchens-100.
\newblock {\em International Journal of Computer Vision}, 130(1):33--55, 2022.

\bibitem{Datasets:OXE}
Abby O’Neill, Abdul Rehman, Abhiram Maddukuri, Abhishek Gupta, Abhishek Padalkar, Abraham Lee, Acorn Pooley, Agrim Gupta, Ajay Mandlekar, Ajinkya Jain, et~al.
\newblock Open x-embodiment: Robotic learning datasets and rt-x models: Open x-embodiment collaboration 0.
\newblock In {\em Proceedings of the IEEE International Conference on Robotics and Automation}, pages 6892--6903, 2024.

\bibitem{robofail}
Zeyi Liu, Arpit Bahety, and Shuran Song.
\newblock Reflect: Summarizing robot experiences for failure explanation and correction.
\newblock {\em arXiv preprint arXiv:2306.15724}, 2023.

\bibitem{egothink}
Sijie Cheng, Zhicheng Guo, Jingwen Wu, Kechen Fang, Peng Li, Huaping Liu, and Yang Liu.
\newblock Egothink: Evaluating first-person perspective thinking capability of vision-language models.
\newblock In {\em Proceedings of the IEEE/CVF Conference on Computer Vision and Pattern Recognition}, pages 14291--14302, 2024.

\bibitem{openeqa}
Arjun Majumdar, Anurag Ajay, Xiaohan Zhang, Pranav Putta, Sriram Yenamandra, Mikael Henaff, Sneha Silwal, Paul Mcvay, Oleksandr Maksymets, Sergio Arnaud, et~al.
\newblock Openeqa: Embodied question answering in the era of foundation models.
\newblock In {\em Proceedings of the IEEE/CVF Conference on Computer Vision and Pattern Recognition}, pages 16488--16498, 2024.

\bibitem{MLLM:GPT4o}
Aaron Hurst, Adam Lerer, Adam~P Goucher, Adam Perelman, Aditya Ramesh, Aidan Clark, AJ~Ostrow, Akila Welihinda, Alan Hayes, Alec Radford, et~al.
\newblock Gpt-4o system card.
\newblock {\em arXiv preprint arXiv:2410.21276}, 2024.

\bibitem{MLLM:Claude3.5Sonnet}
Anthropic.
\newblock Claude sonnet 3.5.
\newblock 2024.

\bibitem{MLLM:LlavaOneVision}
Bo~Li, Yuanhan Zhang, Dong Guo, Renrui Zhang, Feng Li, Hao Zhang, Kaichen Zhang, Peiyuan Zhang, Yanwei Li, Ziwei Liu, et~al.
\newblock Llava-onevision: Easy visual task transfer.
\newblock {\em arXiv preprint arXiv:2408.03326}, 2024.

\end{thebibliography}
